\documentclass[sigconf,10pt]{acmart}
\usepackage{graphicx}
\usepackage{caption}
\usepackage{subcaption}
\usepackage{xspace}
\usepackage{enumitem}
\usepackage{multirow}
\usepackage{hyperref}

\newcommand{\ours}{\textsf{Hydra}\xspace}

\newcommand\rev[1]{{\color{black} #1}}

\begin{document}

\acmYear{2024}\copyrightyear{2024}
\setcopyright{rightsretained}
\acmConference[ACM MobiCom '24]{International Conference On Mobile Computing And Networking}{September 30–October 4, 2024}{Washington D.C., USA}
\acmBooktitle{International Conference On Mobile Computing And Networking (ACM MobiCom '24), November 18--November 22, 2024, Washington D.C., DC, USA}
\acmDOI{10.1145/3636534.3690662}
\acmISBN{979-8-4007-0489-5/24/09.}

\title{\ours: Accurate Multi-Modal Leaf Wetness Sensing \\with mm-Wave and Camera Fusion}

\author{Yimeng Liu$^{1}$, Maolin Gan$^1$, Huaili Zeng$^1$, Li Liu$^2$, Younsuk Dong$^1$, Zhichao Cao$^{1}$}
\affiliation{$^1$Michigan State University\country{USA}  \quad $^2$Tsinghua University \country{China}}

\authornote{Zhichao Cao is the corresponding author.}









\renewcommand{\shortauthors}{Y Liu, M Gan, H Zeng, L Liu, Y Dong, Z Cao}

\begin{abstract}
Leaf Wetness Duration (LWD), the time that water remains on leaf surfaces, is crucial in the development of plant diseases.
Existing LWD detection lacks standardized measurement techniques, and variations across different plant characteristics limit its effectiveness.
Prior research proposes diverse approaches, but they fail to measure real natural leaves directly and lack resilience in various environmental conditions.
This reduces the precision and robustness, revealing a notable practical application and effectiveness gap in real-world agricultural settings.
This paper presents \ours, an innovative approach that integrates millimeter-wave (mmWave) radar with camera technology to detect leaf wetness by determining if there is water on the leaf. 
We can measure the time to determine the LWD based on this detection.
Firstly, we design a Convolutional Neural Network (CNN) to selectively fuse multiple mmWave depth images with an RGB image to generate multiple feature images.
Then, we develop a transformer-based encoder to capture the inherent connection among the multiple feature images to generate a feature map, which is further fed to a classifier for detection.
Moreover, we augment the dataset during training to generalize our model.
Implemented using a frequency-modulated continuous-wave (FMCW) radar within the 76 to 81 GHz band, \ours's performance is meticulously evaluated on plants, demonstrating the potential to classify leaf wetness with up to 96\% accuracy across varying scenarios.
Deploying \ours in the farm, including rainy, dawn, or poorly light nights, it still achieves an accuracy rate of around 90\%.
\end{abstract}
\vspace{-10mm}

\ccsdesc[500]{Computer systems organization~Sensors and actuators}
\ccsdesc[500]{Applied computing~Agriculture}
\keywords{Agricultural IoT, Precision Farming, Multi-Modality Sensing}
\maketitle

\vspace{-4mm}
\section{Introduction}
\label{sec:intro}
\rev{
Agriculture plays a crucial role in the global economy, contributing 4\% to the worldwide GDP and over 25\% in some developing countries~\cite{worldbankAgri}.
However, plant disease's increasing frequency and severity threaten productivity, food security, and biodiversity, particularly in vulnerable regions~\cite{van2015global, Singh2023Climate}.
One critical factor influencing plant disease spread is leaf wetness, which is water on leaf surfaces, occurs due to dew, precipitation, fog, or irrigation and signifies ~\cite{plants12152800, ucscleafwetnes}. 
The duration of the leaf wetness is an essential factor for the growth of pathogens such as Venturia inaequalis~\cite{huber1992modeling, weiss1990leaf}.
Thus, accurate LWD detection is vital for effective disease control~\cite{rowlandson2015reconsidering}.
Numerous agricultural studies have demonstrated that accurate LWD sensing can help protect yields across various crops, such as strawberries~\cite{mackenzie2012use}, grapes~\cite{gubler1999control}, and lettuce~\cite{wu2001validation}.

}

Therefore, researchers have introduced different modalities to enhance the performance of the LWD detection system~\cite{ duvdevani1947optical, gan2023poster, PHYTOS31, resisLWS, nguyen2023bio}.
Nevertheless, existing systems face inherent limitations:

\noindent
\textbf{Wetness detection on real leaf}:
The Leaf Wetness Sensor (LWS) employs synthetic leaves ~\cite{PHYTOS31, resisLWS, nguyen2023bio}, which vary from actual leaves in size, shape, and material, leading to a detection error of up to 30 minutes, as detailed in Section~\ref{sec:overall_performance}. 

\noindent
\textbf{Environment Resilience}: The real environment, dynamic and complex nature due to varying light, wind, and plant diversity, can significantly affect existing detection methods. 
For instance, RGB imaging techniques~\cite{duvdevani1947optical} are highly sensitive to lighting conditions, and mmWave-based approach ~\cite{gan2023poster} struggle with leaf vibrations from the wind.
Additionally, the diverse characteristics of plants, such as shape and size, pose substantial challenges for deploying sensing systems. 

\noindent
\textbf{System efficiency}: 
\rev{
Frequent detection is crucial for accurate disease management, especially on large farms where wetness changes across many plants must be tracked multiple times per hour.
This demand for frequent sensing highlights the need for system efficiency to ensure that more plants can be monitored regularly.
However, compared to traditional cameras and sensors, the mmWave-based approach takes longer to create an RF image by scanning plants~\cite{gan2023poster}.
Therefore, improving system efficiency is essential to maintain high accuracy in disease management.

}

As illustrated in Table~\ref{tab:wetness_detection_technology_compare}, previous works either fail to operate on real leaves~\cite{PHYTOS31, resisLWS, nguyen2023bio} effectively or lack efficiency~\cite{gan2023poster}.
They are also sensitive to environmental changes, resulting in inconsistent performance~\cite{duvdevani1947optical, gan2023poster}.

\begin{figure}[!t]
  \centering
  \includegraphics[width=0.75\linewidth]{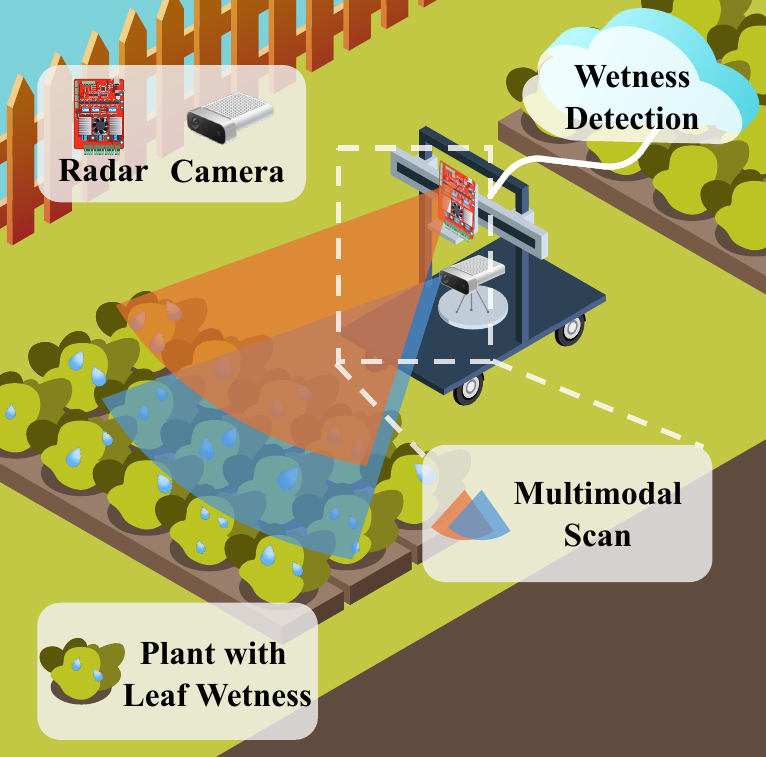}
  \caption{Leaf Wetness detection with \ours.}
  \Description{Wetness detection with \ours}
  \label{fig:system_plant}
  \vspace{-8mm}
\end{figure}

This paper introduces \ours, as shown in Figure \ref{fig:system_plant}, a multi-modal system combining RGB camera and mmWave radar to detect whether the leaf status is wet or dry, achieving high accuracy and efficiency in leaf wetness detection.
The basic observation is that RGB camera and mmWave radar are two orthogonal channels.
RGB images are not sensitive to leaf vibration, and mmWave radar is resilient to various light conditions.
Cameras and mmWave radar generate imaging, which is the input of a Deep Neural Network (DNN) based fusion model.
It includes some layers to enhance its system resilience without model retraining.
To our knowledge, \ours is the pioneering system to combine RGB and mmWave modalities for leaf sensing.
It excels over previous systems in its environmental robustness, high efficiency, and capability for deployment on real leaves.

However, several challenges must be addressed in developing \ours:

\noindent
i) \textbf{How to fusion the mmWave and Camera image:}  
We apply an FMCW chirp, sensitive to varying distances from the mmWave radar, to image the plant.
This technique produces multiple 2D images representing the plant’s cross-sections at different depths, indicating the distance from the cross-sections to the radar.
Meanwhile, RGB cameras generate a singular image that captures surface details influenced by lighting conditions. 
Developing a method that effectively combines a mmWave image at a certain depth with the high-resolution surface detail from RGB images is crucial for creating a comprehensive view of plant wetness.
However, due to the different ways of representing leaf spatial information, it is challenging to fuse mmWave and camera to utilize their orthogonal information. 
To address this challenge, we design a single-depth modality fusion and feature extraction method that takes multiple mmWave images at different depths and an RGB image as inputs.
Specifically, for each depth, we selectively mask the mmWave image at that depth with the RGB image to generate a fusion depth image.
Then, we design a CNN model to extract multiple features from these fusion depth images at different depths.

\noindent
ii) \textbf{How to combine the fusion features at a 3D level:} Integrating the fusion features at different depths necessitates the development of a robust 3D classification model capable of accurately interpreting complex, multi-dimensional information. 
However, since the 3D structure of various crops is usually complex and challenging to model, finding a way to connect the features at different depths as a representative feature map is not trivial.
To solve this problem, we develop a transformer-based feature encoder, which treats the fusion features at different depths as a serial sequence and automatically finds the inherent 3D domain connection among them to generate an informative feature map.

\begin{table}[!t]\footnotesize
\caption{Comparison Methods for Detecting LWD}
  \begin{tabular}{c|ccc}
    \toprule
    \multirow{2}{*}{Methods} & Real & System & Environment \\
    & Leaf & Efficiency & Resilience \\
    \midrule
    \texttt{LWS~\cite{PHYTOS31, resisLWS, nguyen2023bio}} & No & Yes & No \\
    \texttt{Infrared/RGB cameras~\cite{duvdevani1947optical}} & Yes & Yes & No \\
    \texttt{mmWave radar~\cite{gan2023poster}} & Yes & No & No \\
    \textbf{\texttt{\ours}} & \textbf{Yes} & \textbf{Yes} & \textbf{Yes} \\
    \bottomrule
  \end{tabular}
  \label{tab:wetness_detection_technology_compare}
    \vspace{-6mm}
\end{table}

\noindent
iii) \textbf{How to increase the system efficiency of the mmWave imaging and keep the high accuracy:}  
mmWave technology produces high-resolution images through its short wavelength and SAR technique, proving essential for detailed observations.
However, the scanning period significantly degrades the system efficiency. 
Enhancing system efficiency while preserving high-resolution imaging in mmWave technology is complex due to the compromise between synthetic aperture size and detail accuracy.
Finding ways to boost efficiency without sacrificing the quality of leaf wetness detection remains a significant challenge.
To tackle this challenge, we observe that our two-modality fusion enhances the feature domain, allowing us to shrink mmWave Radar's Field-of-View (FoV).
We design an adaptive FoV adjustment method to balance the system's efficiency and accuracy.

\noindent
iv) \textbf{How to adapt the system in different scenarios:} 
In real-world conditions, plants are subjected to constantly changing environments that can significantly impact the performance of leaf wetness detection.
Therefore, it is imperative to enhance \ours's robustness, ensuring it can accurately detect and analyze leaf wetness under a wide range of environmental conditions.
This adaptation involves creating algorithms resilient to light, temperature, and movement changes, ensuring reliable detection across diverse agricultural settings.
Addressing these challenges is vital for advancing precision agriculture and contributing to more effective plant disease management and crop production strategies.
To address this challenge, we augment the datasets used for model training. 
Specifically, we include different lighting conditions for camera images.
For mmWave radar, we manually generate diverse leaf vibration patterns with a fan for different leaf types, sizes, gestures, etc.

We implement \ours with commercial off-the-shelf hardware components.
By applying our model to various plants in different environmental scenarios, we have successfully trained a deep-learning model notable for its accuracy and robustness.
Extensive testing demonstrates that \ours attains an outstanding accuracy rate of 96.63\% under indoor conditions with simulated scenarios and maintains a performance level of about 90\% in the farm, including rainy conditions, dawn, and poorly nighttime settings.
Moreover, \ours exhibits exceptional performance by reducing the error margin in LWD detection to approximately 2 minutes, markedly outperforming previous systems.
 
The contributions of this paper can be summarized as follows:
\begin{itemize}[topsep=0pt, partopsep=0pt]
    \item To our knowledge, \ours is the first contactless, multi-modality sensing system designed explicitly for LWD detection.
    It can directly scan plant surfaces for accurate, robust, and efficient wetness identification.
    \item We have incorporated novel multi-modality techniques,  leveraging the synergistic capabilities of mmWave imaging and camera image that allow for an advanced data fusion, significantly mitigating the challenges of aligning disparate dimensional data.
    \item We improved the SAR system's efficiency by 25\%, enabling precise leaf wetness detection.
    Combined with RGB imaging, this enhancement allows for accurate assessments within a limited field of view and using a smaller synthetic aperture.
    This optimization notably enhances the system's ability to measure moisture levels accurately.
    \item The system demonstrates robust performance in complex outdoor scenarios, maintaining high efficiency even when specific modalities may not function optimally.
    When our model is applied to the farm, it achieves around 90\% accuracy, showcasing its adaptability and effectiveness in varied environmental settings.

\end{itemize}
\section{Preliminary and Motivation}
In this section, we introduce the definition of Leaf Wetness and Leaf Wetness Duration and review prior efforts in LWD.
We begin with the existing LWS and the RGB camera solution. 
We then discuss approaches that utilize mmWave radar and outline the motivation behind our research.

\begin{figure}[!t]
   \centering
   \begin{subfigure}[b]{0.185\textwidth}
       \centering\includegraphics[width=\textwidth]{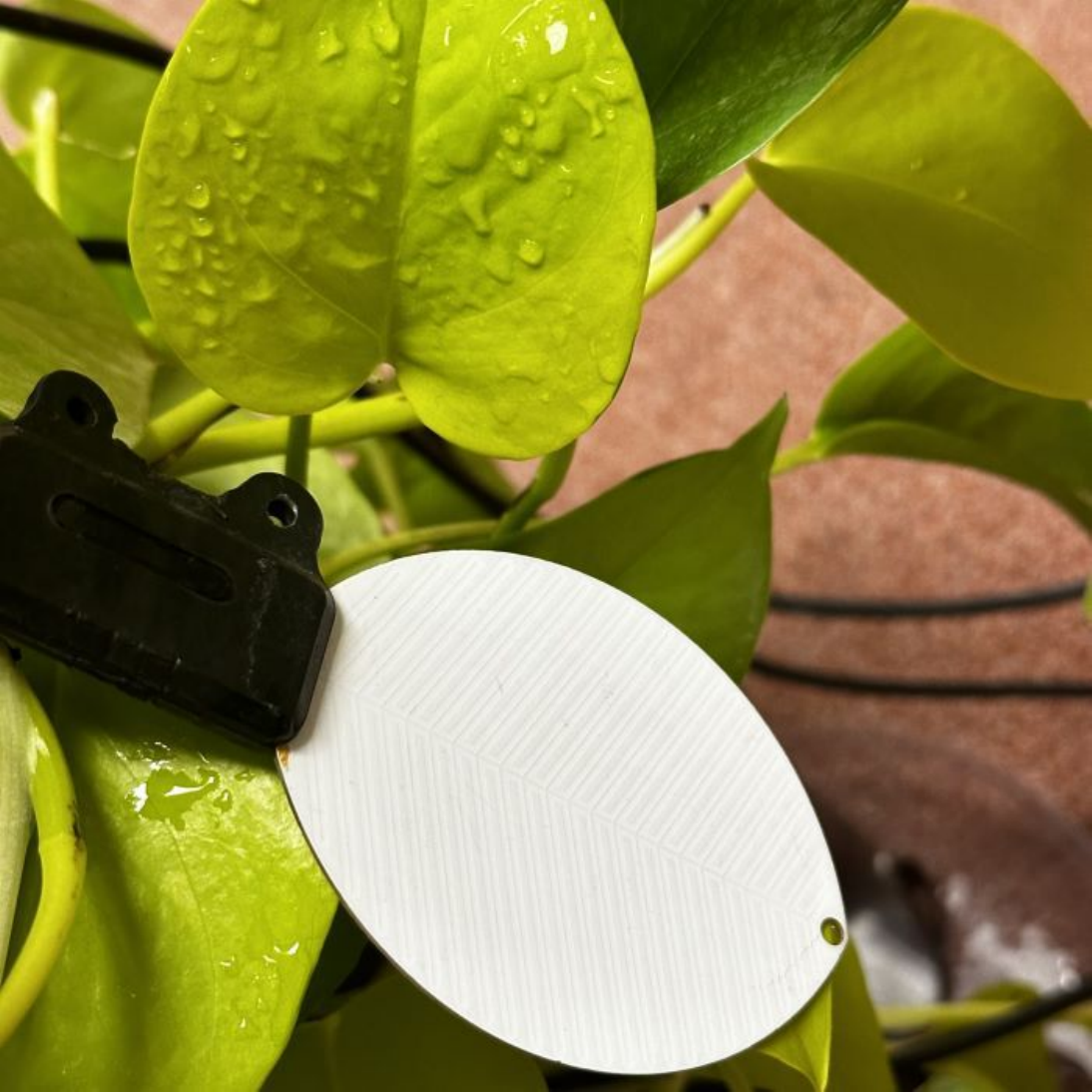}
       \caption{LWS Error Scenario}
    \label{fig:lws-error}
   \end{subfigure}
   \begin{subfigure}[b]{0.255\textwidth}
       \centering\includegraphics[width=\textwidth]{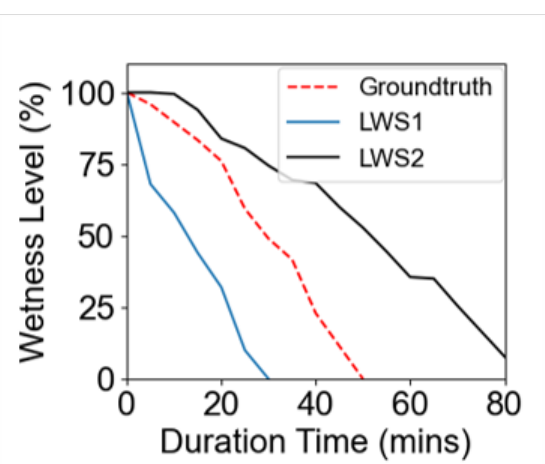} 
       \caption{LWS Performance}
        \label{fig:lws-analysis}
   \end{subfigure}
    \caption{LWS Anlaysis. Leaf wetness error scenario demonstrating the inaccuracy of LWS~\cite{PHYTOS31}. }
    \Description{Setup of the LWS and the performance skew from the ground truth}
    \vspace{-5mm}
\end{figure}

\rev{
\subsection{Definition of Leaf Wetness}
Leaf Wetness Duration is the period during which a leaf remains wet.
Accurately detecting LWD is crucial for understanding and managing plant diseases ~\cite{rowlandson2015reconsidering}.
A vital part of determining LWD is accurately identifying Leaf Wetness, which refers to the presence of water on a leaf's surface~\cite{plants12152800, ucscleafwetnes}.
This condition can be classified into two states: dry or wet.
In our study, we obtain the ground-truth leaf wetness using a moisture meter specified in Section \ref{sec:data-collect}.
We calibrate the meter in a dry, ventilated indoor environment to establish a baseline threshold.
If the moisture reading exceeds this threshold, the leaf is considered wet. Otherwise, the leaf is dry.
}

\subsection{Leaf Wetness Sensor}
\label{sec:lws-preliminary}
The Leaf Wetness Sensor ~\cite{PHYTOS31, resisLWS, nguyen2023bio} mimics the leaf's shape and positioning to simulate the real leaf status.
Within this domain, LWSs are categorized into resistive, using hydrophilic to detect wetness through resistance changes \cite{resisLWS}, and capacitive types, leveraging dielectric material coatingxs to detect wetness via capacitance shifts, based on the differing dielectric constants of water and air ~\cite{PHYTOS31}.
\rev{The Bio-Mimetic LWS \cite{nguyen2023bio} uses the capacitive LWS to simulate the surface and texture of leaves to detect LWD more accurately.
These LWSs can represent the change of the free water on the leaf.
However, a major issue with these sensors is that they do not directly measure the wetness of actual leaves, so they may not precisely show the result.}

\rev{We conduct an experiment using two different Capacitive LWS \cite{PHYTOS31} placed at various locations on the plant to detect changes in wetness.
We record the wetness level by normalizing the LWS readings to a 0-100 scale, in which 0 represents complete dryness and 100 represents saturated wetness.
The ground truth detection is based on the wetness meter described in Section \ref{sec:data-collect}. 
We record and normalize the wetness meter readings to a 0-100 scale.}
Our observation in Figure \ref{fig:lws-error} indicates that the sensors typically weigh more than actual leaves, causing them to slide more easily and leading to quicker water runoff than real leaves.
Additionally, in specific scenarios, the sensor's edge texture may trap water, preventing evaporation and resulting in prolonged wetness duration compared to natural leaf surfaces. 
We conduct an empirical study to illustrate the difference in wetness level between the ground truth and LWSs in the same environment. 
The analysis in Figure \ref{fig:lws-analysis} reveals that the sensors varying placements on the plant lead to inconsistent results, significantly deviating from the established ground truth.
This limitation underscores the need for higher fidelity in LWD detection and contributes to more precise and reliable agricultural monitoring systems.

\begin{table}[!t]\footnotesize
\caption{Camera Performance in Different Lighting}
  \label{tab:camera-comparison}
  \begin{tabular}{c|ccc}
    \toprule
Lighting Condition & Normal Light & Dawn & Evening \\
    \midrule
\texttt{Camera Accuracy} & 90.42\% & 72.73\% & 68.97\% \\
    \bottomrule
  \end{tabular}
\end{table}

\subsection{RGB Camera Recognition}
\label{sec:camera-preliminary}
\rev{
Exploring LWD through RGB cameras offers clear distinctions between dry and wet leaves, with obvious features with drops on the leaf that can be effectively detected with DNN under ideal lighting conditions ~\cite{patel2021strawberry}.}
However, the accuracy significantly drops in agriculture, where lighting varies—during dawn, night, or cloudy days.
The performance of only a camera for detection across various lighting conditions decreased.
We empirically study the influence of light conditions on RGB image-based detection.
We are using ResNet-18 ~\cite{ResNet2016} to classify if the leaf is dry or wet with the RGB image.
The results are summarized in Table~\ref{tab:camera-comparison}.
The result in normal light is around $90\%$, but when the lighting condition degrades, it drops to $70\%$. 
Also, RGB cameras rely on reflected light, limiting their ability to detect conditions behind or beneath overlapping leaves.
They cannot penetrate the plant's canopy to reveal hidden wetness.

\subsection{mmWave-based Sensing}
\label{sec:mmWave-preliminary}
mmWave, operating at the higher frequencies of the electromagnetic spectrum, leverages the advantage of large bandwidth at shorter distances ~\cite{wikner2011progress}. This characteristic limits penetration and propagation depth in underwater environments, as high-frequency electromagnetic (EM) signals have limited penetration depth~\cite{Qureshi2016}. Conversely, this limitation benefits surface texture applications \cite{Liang2021water}. The precision and sensitivity afforded by mmWave technology, mainly through the dramatic drop of the reflection, enable the detection of minute wetness variations on leaf surfaces. 

\rev{Synthetic Aperture Radar (SAR) is widely used in radar applications for its ability to produce high-resolution images by simulating a large aperture through the relative motion between the radar and the target.
A recent study demonstrated the use of SAR in enhancing imaging resolution and environmental perception for autonomous systems in real-time conditions~\cite{zhou2009applications}.
The increased aperture size allows SAR to capture fine-grained images, revealing intricate details that improve leaf wetness detection and enable precise monitoring and accurate identification of leaf status.
Using the FMCW chirp, mmWave radar is sensitive to the distance between the radar and the target.
SAR can image targets at different depths, providing cross-sectional information at varying distances from the radar.
}

\begin{table}[!t]\footnotesize
\caption{SAR Image Performance in different scenario.}
  \label{tab:sar-comparison}
  \begin{tabular}{c|cc}
    \toprule
    Wind Condition& No Wind & Moderated Wind \\
    \midrule
    \texttt{SAR Image Accuracy} & 90.85\% &  70.43\% \\
    \bottomrule
  \end{tabular}
\end{table}

\begin{figure}[!t]
   \centering
   \begin{subfigure}[b]{0.14\textwidth}
       \centering\includegraphics[width=\textwidth]{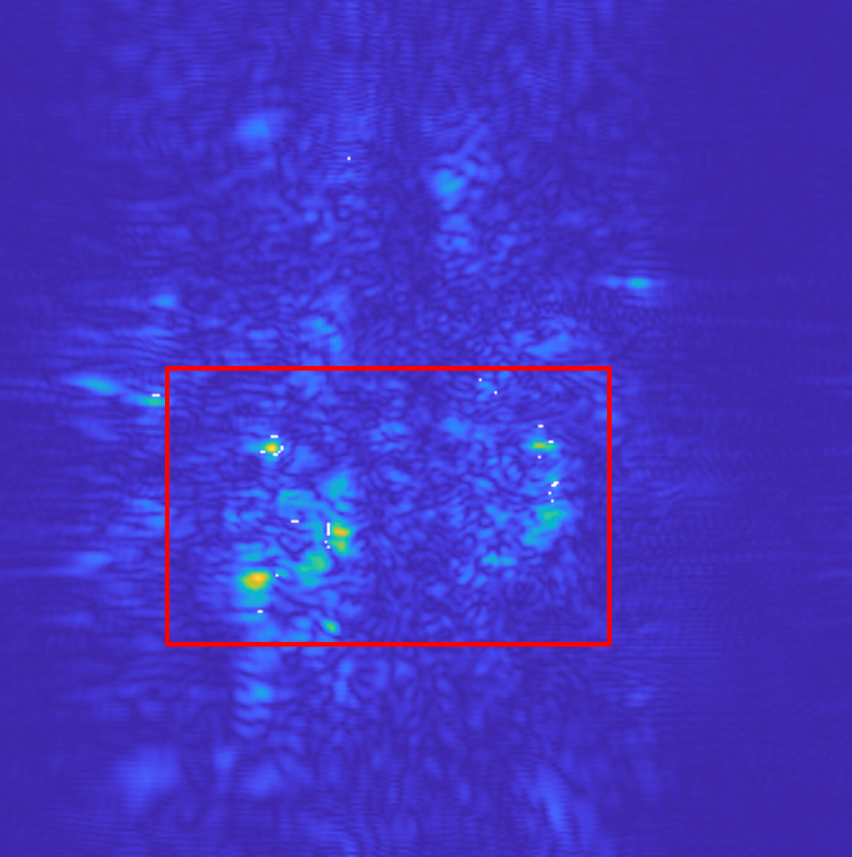} \\
       \hfill
       \caption{SAR Imaging}
       \label{fig:sar-example}
   \end{subfigure}
   \hfill
   \begin{subfigure}[b]{0.33\textwidth}
       \centering\includegraphics[width=\textwidth]{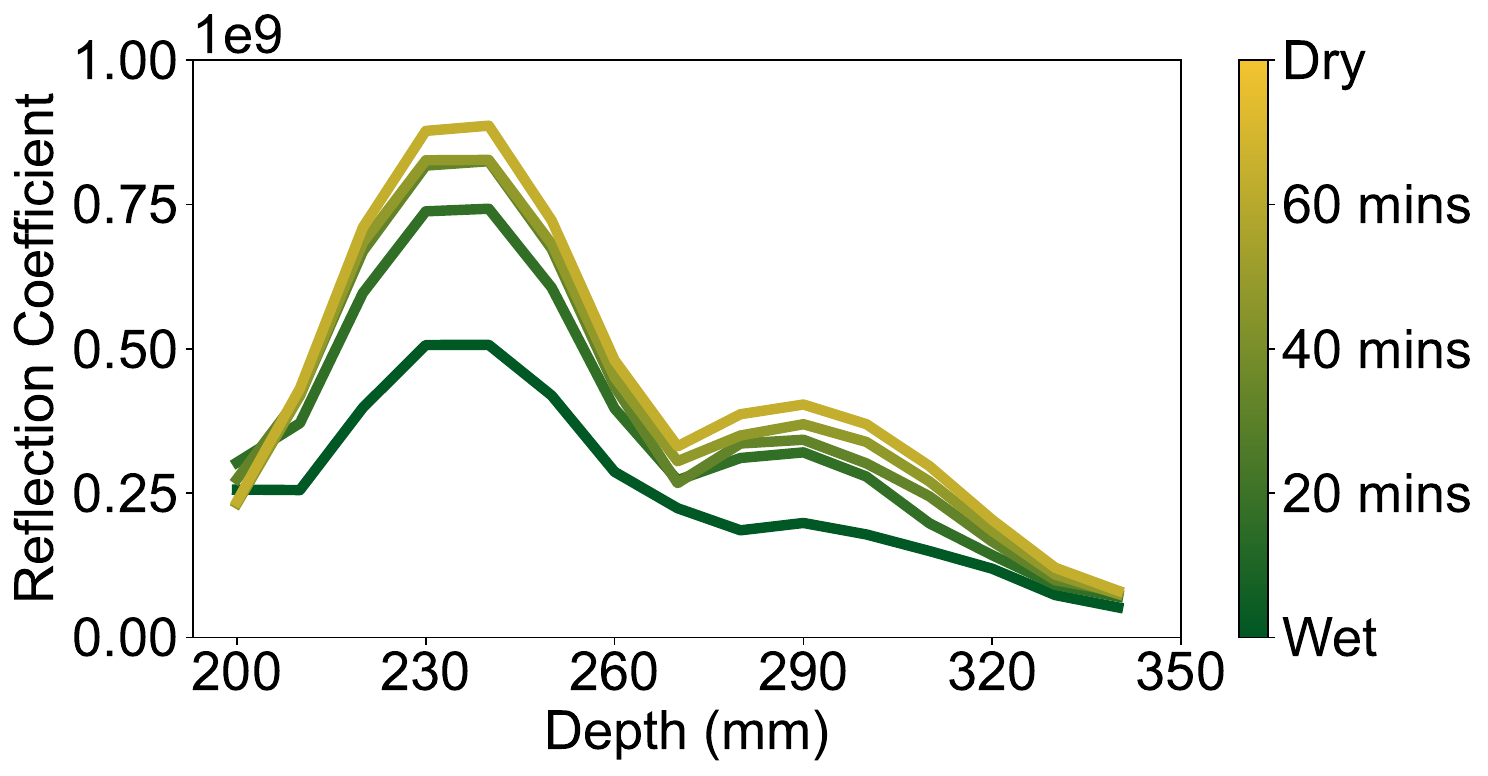} 
       \caption{Wetness Stage vs Reflection Coefficient}
        \label{fig:mmWave_duration_reflection}
   \end{subfigure}
   \caption{Example of SAR imaging results and the signal reflection change as the leaf dries.}
   \Description{SAR Imaging example and the relationship with wetness level}
   \vspace{-4mm}
\end{figure}

From the methodology described by ~\cite{yanik2020development}, we use a testbed incorporating the mmWave to apply the SAR technique imaging a potted plant, capturing the reflected signals from the leaves.
This approach detects leaf wetness by directly targeting the plant, offering a sensitive solution that avoids mere sampling. 
In our adaptation of ~\cite{gan2023poster}, we position a stationary plant approximately 200 mm from the radar, monitoring the transition from wet to dry conditions. 
Observations are recorded at $20$-minute intervals, with the plant reaching complete dryness after 80 minutes.
Figure~\ref{fig:sar-example} illustrates an example of a SAR imaging result in 230mm depth, a cross-section of a plant 230 mm away from the radar.
The image features an area highlighted in yellow, encircled by a red rectangle, designating the leaf surface information detected by the radar.
The irregular shape of plants makes it difficult to identify the specific part of the plant depicted in an image and to detect features related to wetness.
The analysis shown in Figure \ref{fig:mmWave_duration_reflection} suggests that within a depth range of 200-340mm, the leaf surface area reflection signals from the plant increase as it dries.
This observation highlights the potential of using reflection intensity as a feature of plant dryness, offering insights into plants' wetness content.

Current methods have notable limitations, especially in terms of their robustness. 
The accuracy predominantly depends on analyzing data from a constrained depth range, which inadequately addresses leaf wetness and plant health comprehensively.
Apart from that, the environment poses a significant challenge for mmWave imaging.
We use the ResNet-18 model to classify the SAR image between dry and wet.
The results in Table \ref{tab:sar-comparison} show that wind significantly impacts SAR imaging accuracy. 
In conditions with no wind, the accuracy can reach 90\%, but with moderate wind, the precision drastically drops to 70\%.
This finding highlights the challenges of SAR imaging in windy conditions.

\subsection{Motivation}
We have observed that no existing single-modality leaf wetness sensing techniques can achieve high accuracy under various environmental conditions.
This motivates us to consider a multi-modality solution to enable high-accuracy leaf wetness sensing with strong environmental resilience while keeping system efficiency as high as possible.

\section{System Overview}
\begin{figure*}[ht]
  \centering
  \includegraphics[width=0.9\linewidth]{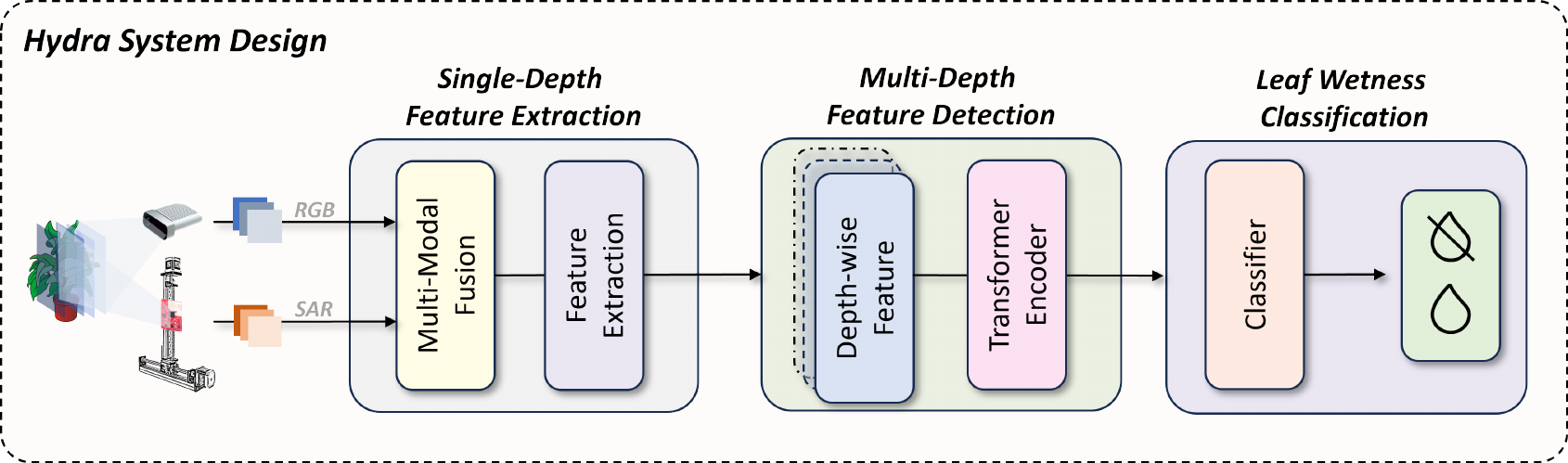}
  \caption{\ours Overview. \ours contains three main procedures: Single-Depth Feature Extraction, Multi-Depth Feature Detection, and classifier. We fuse the multi-modality and extract wetness features in the Single-Depth Feature Extraction phase. The Multi-Depth Feature Detection stage leverages these features at a 3D level, culminating in utilizing a leaf wetness classification algorithm to derive the final wetness assessment.}
  \label{fig:system_overview}
    \Description{Overview of the \ours design include all parts}
    \end{figure*}

We present \ours, an advanced multi-modal leaf wetness detection system. \ours integrates mmWave and RGB imaging to facilitate water on the leaf wetness detection across various plant species.
Then, we use the detection result to calculate LWD to help with disease control.
Its precise and durable system architecture, shown in Figure ~\ref{fig:system_overview}, consists of the main components: SAR Imaging System, Single-Depth Feature Extraction, Multi-Depth Feature Detection, Model Training and Data Enhancement.

\textbf{SAR Imaging System.} In this module, we design the SAR imaging system that transmits mmWave signals and senses the target plants from various depths. Our approach leverages the foundational design principles outlined in ~\cite{gan2023poster} to optimize efficiency by incorporating an RGB camera.

\textbf{Single-Depth Feature Extraction.} In this module, \ours system introduces a groundbreaking fusion method that integrates data from mmWave radar images captured in a certain depth with RGB camera images containing surface information.
By merging these diverse modalities with different ways of representing leaf spatial information, we generate a comprehensive input set that enhances the accuracy of subsequent detection processes.

\textbf{Multi-Depth Feature Detection.} In this module, our approach employs an innovative fusion method that consolidates data across multiple depths for the final classification outcome.
This strategy develops a thorough insight into the plant's characteristics, culminating in a holistic understanding that informs the overall result.

\textbf{Model Training and Data Enhancement.} This module covers our strategies for assessing and improving the model's performance under various environmental conditions, aimed at increasing its robustness and versatility for complex agricultural scenarios.

\section{\ours Design}
This section explores the methodologies behind \ours, beginning with an overview of SAR imaging techniques and proceeding to the extraction of single-depth features. It then delves into the detection of leaf wetness using multi-depth feature analysis. Finally, we discuss the model training protocols and data enhancement strategies employed.
\subsection{Variable Scan Distance SAR Imaging}
In this section, we develop our imaging system based on SAR technology. 
This system emits signals at regular intervals to capture images of relatively stationary targets.
One key strength of SAR technology is its ability to increase the aperture size effectively. 
With the combination of FMCW and SAR, imaging technology has the advantages of being lightweight and low-cost and is widely used in near-field imaging scenarios~\cite{meta2007signal,stove1992linear,yanik2019near,yanik2020development}. 

For the FMCW signal, the instantaneous frequency can be expressed as a linear function as
\begin{equation}
    m(t) = \cos[2\pi(f_0t + 0.5Kt^2)]
\end{equation}
where \(f_0\) represents the carrier frequency at the initial time \(t=0\), and \(K = B/T\) denotes the frequency modulation slope, derived from the sweep bandwidth \(B\) and the chirp duration \(T\). 
Upon receiving the backscattered signal, the radar system employs a dechirping process that isolates the beat frequency by mixing the received signal with its in-phase (\(s_I(t)\)) and quadrature (\(s_Q(t)\)) components. 
This yields a complex beat signal expressed as
\begin{equation}
    s(t) = s_I(t) - js_Q(t) = \sigma e^{-j2\pi(f_0\tau + K \tau t -0.5 K\tau^2)},
\end{equation}
with \(\tau\) signifying the round-trip delay of the echo, and \(\sigma\) encompassing both the target's reflectivity and amplitude decay.

For the spatial representation of the received signal, particularly in the wave number domain, the signal can be simplified as
\begin{equation}
s(x', y_T, y_R, k) = \iiint p(x, y, z) e^{-jkR_T} e^{-jkR_R} \, dx \, dy \, dz,
\end{equation}
where \(R_T\) and \(R_R\) denote the distances from the transmitter and receiver to the scatter point, respectively, aligning within a Cartesian coordinate system where \(x\), \(y\), and \(z\) axes define the scanning, vertical, and depth dimensions.

To accurately model the distances \(R_T\) and \(R_R\) about the transmitter and receiver's positions, we consider
\begin{equation}
    \begin{split}
    R_T = \sqrt{(x + (x' + \delta_T/2))^2 + (y-y_T)^2 + (z-Z_0)^2},\\
    R_R = \sqrt{(x - (x' + \delta_T/2))^2 + (y-y_T)^2 + (z-Z_0)^2}.
    \end{split}
\end{equation}
This geometric consideration is vital for the subsequent image reconstruction phase, where the range migration algorithm is adapted for the SAR context.
The conversion from multi-static to a monostatic equivalent involves phase compensation, simplifying the data processing for precise depth-based imaging as
\begin{equation}
    \widetilde{s}(x',y') = \iint p(x, y)e^{-j2kR}dx dy,
\end{equation}
highlighting the distance \(R\) to the scatter point and facilitating the extraction of 2D images from the radar data.

Finally, incorporating Weyl's representation theorem ~\cite{weyl1919ausbreitung} enables the approximation of spherical waves through plane waves, leading to the backscatter data being expressed as
\begin{equation}
    \widetilde{S}(k_x, k_y) = P(k_x, k_y)e^{jk_zZ_0},
\end{equation}
where the inverse Fourier transform applied to \(\widetilde{S}(k_x, k_y)\) yields the reconstructed 2D image of the target area to get the 2D image:
\begin{equation}
    p(x, y) = IFT^{(k_x, k_y, k_z)}[e^{-jk_zZ_0}\widetilde{S}(k_x, k_y)],
\end{equation}
Adjusting the $x$ range balances a broader field-of-view and higher resolution versus a shorter view for enhanced efficiency.
This enables flexibility in capturing detailed imagery or prioritizing rapid scanning based on specific requirements.

\subsection{Single-Depth Feature Extraction}
\label{sec:single-depth-fe}
Integrating SAR imaging with camera images poses a significant challenge due to their different ways of representing leaf spatial information.
Enhanced by FMCW chirps, SAR imaging excels at detecting distance variations, allowing for the extraction of detailed features at multiple depths.
This capability includes identifying features not visible to the RGB camera.
Conversely, camera imaging relies on light reflection to capture information, project, and record data based on light reflected off the scanning plane.
This method can capture details from multiple depth layers within a single image.
This difference complicates the direct integration of the two image types, as they provide information in distinct dimensions.

\begin{figure}[!t]
  \centering
  \includegraphics[width=\linewidth]{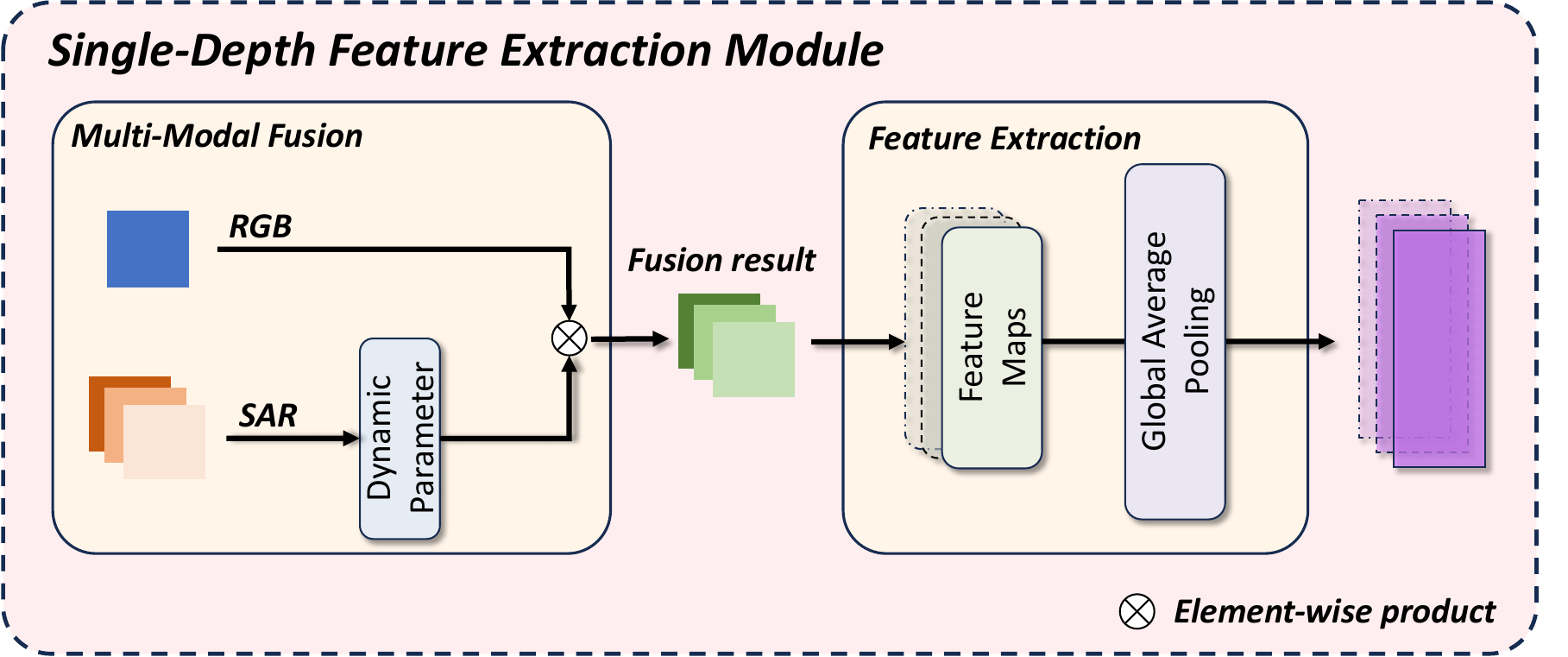}
  \caption{Single-Depth Feature Extraction architecture.}
  \label{fig:single_depth}
   \Description{Single-Depth module}
\end{figure}

To tackle this challenge, we introduce a novel approach incorporating a machine learning layer designed to fuse these two modalities, as illustrated in Figure \ref{fig:single_depth}.
We employ the concept of a mask for dimension alignment, using a certain depth SAR image to mask the camera image.
We've adapted this conventional computer vision masking technique to spotlight regions of interest. 
We normalize the SAR image to a $[0, 1]$ scale, where higher values indicate pixels of greater relevance.
This normalized mask is then applied to the camera image to extract the information from the depth specified by the SAR image.
Moreover, the fusion process preserves the SAR image's ability to reveal surface textures through reflection rates.
We also introduce a dynamic parameter within the layer to adjust the balance between the SAR and camera images based on feature correspondence.

Following the data fusion, the feature detection process becomes significantly more complex.
In this context, CNNs emerge as an invaluable tool for our analysis.
CNNs excel in navigating the complexities of high-dimensional data, adeptly isolating and enhancing features critical for accurate classification.
The approach of mimicking the hierarchical pattern recognition found in the human visual system enables the layered structure to progressively refine inputs, starting with essential edge detection and advancing to more complex feature identification~\cite{cnnBrain2023St-Yves}.

\begin{figure}[!t]
  \centering
  \includegraphics[width=0.85\linewidth]{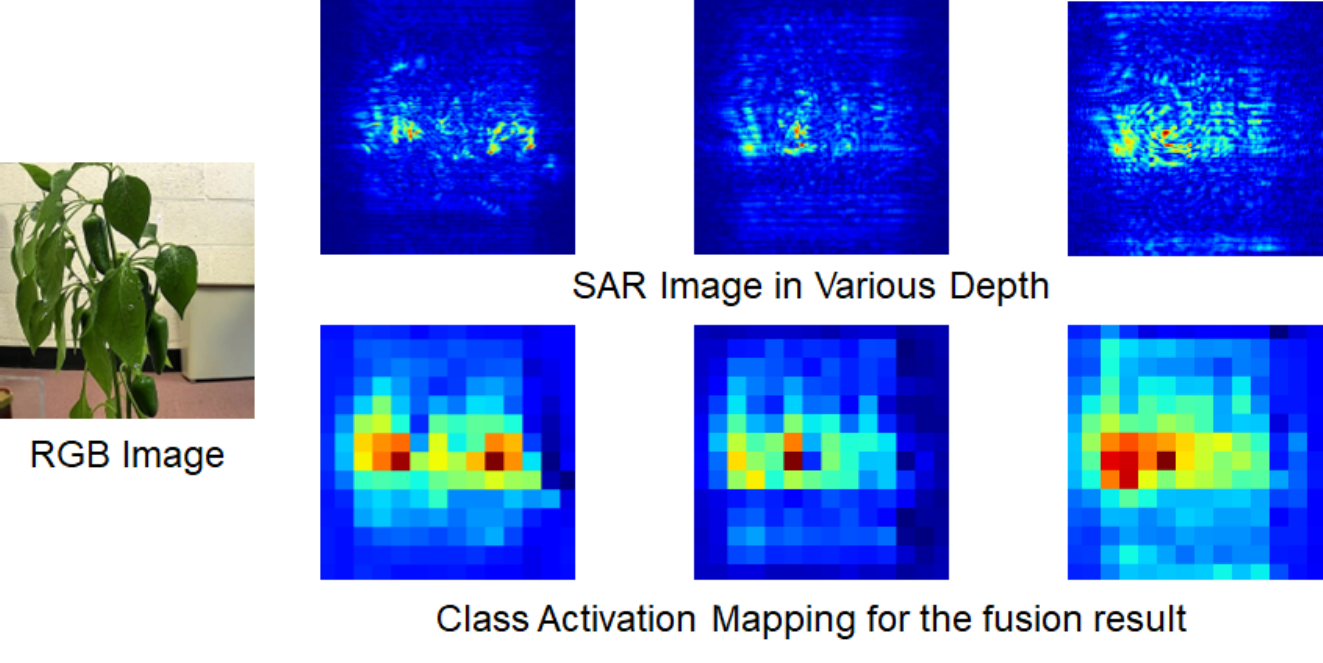}
   \caption{Example of Class Activation Mapping Result.}
   \label{fig:CAM_Result}
   \Description{Single-Depth module}
\vspace{-2mm}
\end{figure}

In the concluding phase of our Single-Depth Feature Extraction module, we introduce a Global Average Pooling (GAP) layer~\cite{lin2014network}.
This layer calculates the spatial mean of feature maps from the terminal convolutional layer, streamlining data complexity while preserving critical feature information without adding new trainable parameters.
The GAP layer is a structural regularize.
It reduces overfitting by simplifying the architecture and improving the model's interpretability.
It can enable the application of Class Activation Mapping (CAM) to produce heatmaps to show the model's focus areas.
This feature is handy when analyzing SAR images and the outputs from the multi-modality fusion module, where direct feature visualization is challenging. From Figure \ref{fig:CAM_Result}, we can get the example of RGB image, SAR image in various depths, and CAM.
Following the fusion of multiple modalities, this example shows that the CAM heatmap reveals the model's focus on leaf areas vital for wetness evaluation, corroborated by both the camera and SAR imagery.

 \rev{
To enhance the model's receptivity to detailed SAR features, we selectively exclude camera data from the fusion process and focus solely on SAR imaging.
We de-emphasized RGB imagery in training by randomly removing 20\% of the images from the dataset.
This approach encourages the model to develop a deeper understanding of SAR features and achieve a well-rounded grasp of multi-modal data.
By employing this method, we significantly boosted the model's ability to identify subtle wetness cues on leaves.
 }

\subsection{Multi-Depth Leaf Wetness Detection}
\label{sec:multi-depth}
\begin{figure}[t]
  \centering
  \includegraphics[width=\linewidth]{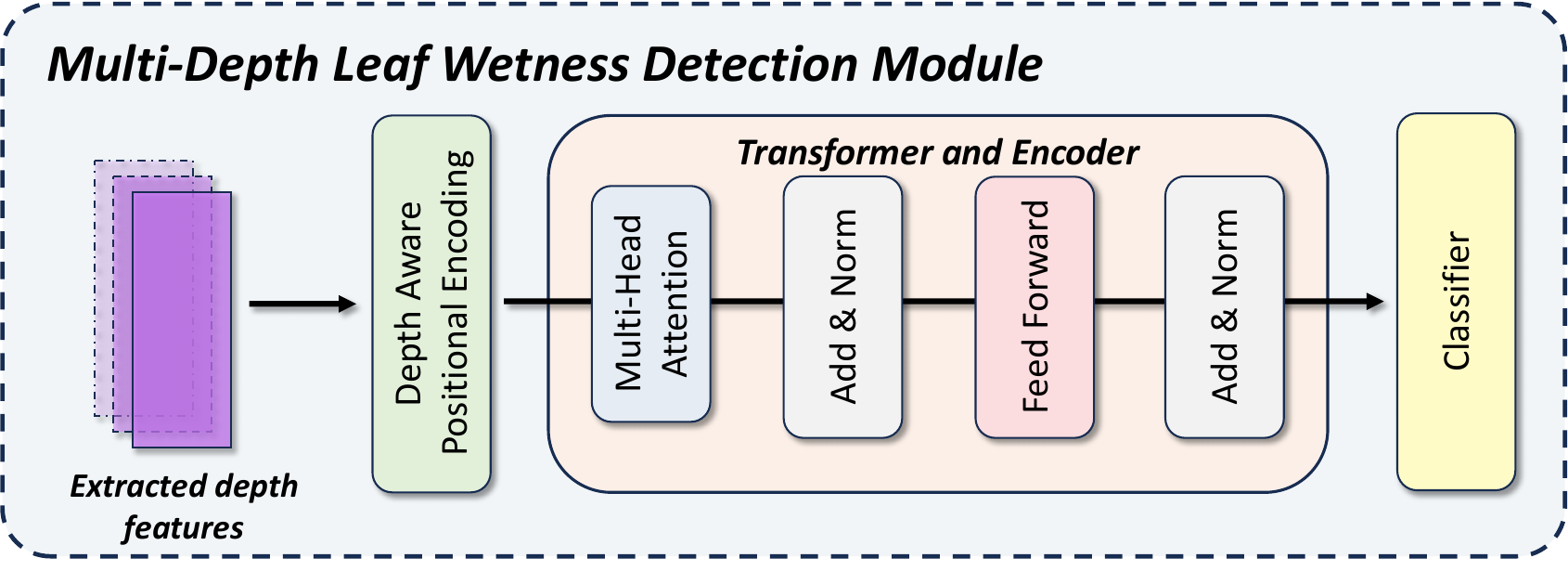}
  \caption{The Multi-Depth leaf wetness Detection module's architecture.}
  \label{fig:multi-depth}
  
   \Description{Multi-Depth Module}
\vspace{-4mm}
\end{figure}

\rev{A notable challenge in plant analysis is the complexity of plant structures.
Varying leaf sizes, orientations, and densities make accurate interpretation difficult.
While 3D models provide a detailed view and can improve the accuracy of the analysis.
}
We have developed a model by integrating and classifying data across multiple depths, enabling practical 3D analysis shown in Figure \ref{fig:multi-depth}. We first introduce depth-aware positional encoding ~\cite{gehring2017positionawareness}, which is meticulously designed to incorporate depth positional information into our model and help spatial positioning across depth dimensions. The mathematical underpinning of this layer is expressed through the positional encoding formula:

\begin{equation}
     PE(depth, i) = \begin{cases} 
    \sin\left(\frac{depth}{10000^{2i/d_{model}}}\right) & \text{for even } i \\
    \cos\left(\frac{depth}{10000^{2(i-1)/d_{model}}}\right) & \text{for odd } i
    \end{cases} 
\end{equation}

Here, \(depth\) signifies the position within the depth sequence, \(i\) represents the dimension within the feature space, and \(d_{model}\) indicates the feature space dimensionality.
Then, we implement a multi-head attention mechanism reminiscent of those in cutting-edge Large Language Models (LLMs), which endows our transformer encoder with exceptional contextual awareness \cite{chatbot2023c, dynamic2023c, llm2023c}.
This feature supports comprehensive analysis throughout the plant by merging data from various depths, offering a detailed and precise assessment of leaf.
The multi-head attention mechanism concurrently focuses on multiple depth-related aspects to pinpoint subtle moisture shifts.
Integrating depth-aware positional encoding and multi-head attention mechanism improves our model's precision in leaf wetness detection by addressing the complexities of spatial analysis and positional data synthesis.

\subsection{Model Training and Data Enhancement}
\label{sec:model_training}
Training the model for leaf wetness detection involves a meticulously designed process to accurately classify leaves' wetness and dryness states.
The foundation of our model's training regimen is based on utilizing binary cross-entropy as the loss function.
This decision is based on the effectiveness of binary cross-entropy for binary classification tasks, which makes it especially apt for differentiating between the binary states of leaf surface conditions.
The loss function can be shown as
\begin{equation}
L(y, \hat{y}) = -\frac{1}{N} \sum_{i=1}^{N} [y_i \log(\hat{y}_i) + (1 - y_i) \log(1 - \hat{y}_i)],
\end{equation}
where \(L\) is the loss, \(N\) is the number of observations, \(y_i\) represents the actual label for the \(i^{th}\) observation, and \(\hat{y}_i\) denotes the predicted probability of the leaf being wet.
Our model's training begins with the pretraining phase of the Single Depth Fusion Module, an essential element for accurately detecting leaf wetness across varying depths. We will apply Section \ref{sec:multi-depth} with the extracted feature to combine the feature analysis and produce the final classification results.
Moreover, we introduce simulations of breeze-induced motion, a condition under which SAR imaging may not faithfully capture the scene.
We train our model to recognize and adjust to such anomalies.
This holistic data augmentation strategy, addressing multiple real-world scenarios, significantly enhances the model's resilience and relevance to environmental conditions.

\section{Implementation}

\subsection{System Implementation}
\begin{figure*}[!t]
   \centering
   \begin{subfigure}[b]{0.196\textwidth}
       \centering
       \includegraphics[width=\textwidth]{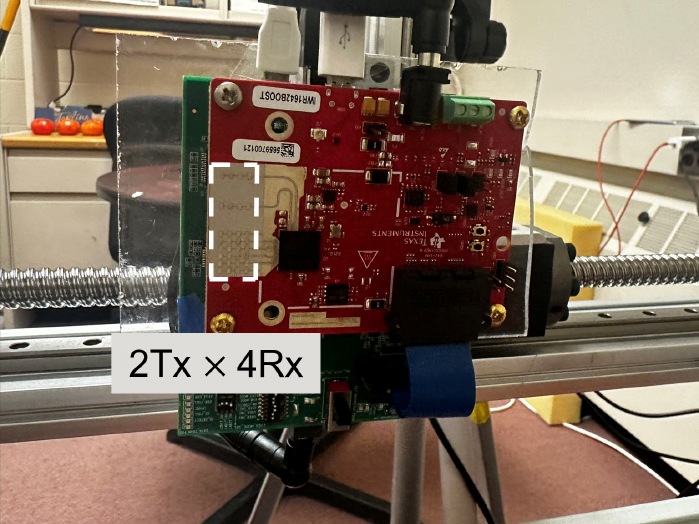}
       \caption{mmWave Radar}
   \end{subfigure}
   \begin{subfigure}[b]{0.196\textwidth}
       \centering
       \includegraphics[width=\textwidth]{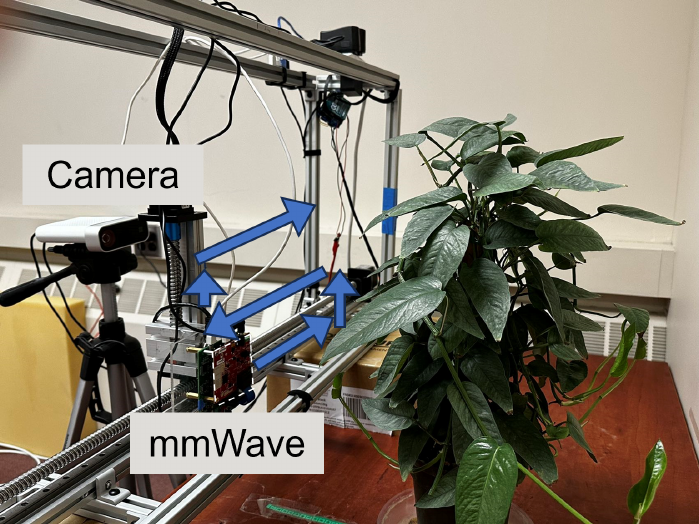}
       \caption{Indoor Setup}
   \end{subfigure}
   \begin{subfigure}[b]{0.196\textwidth}
       \centering
       \includegraphics[width=\textwidth]{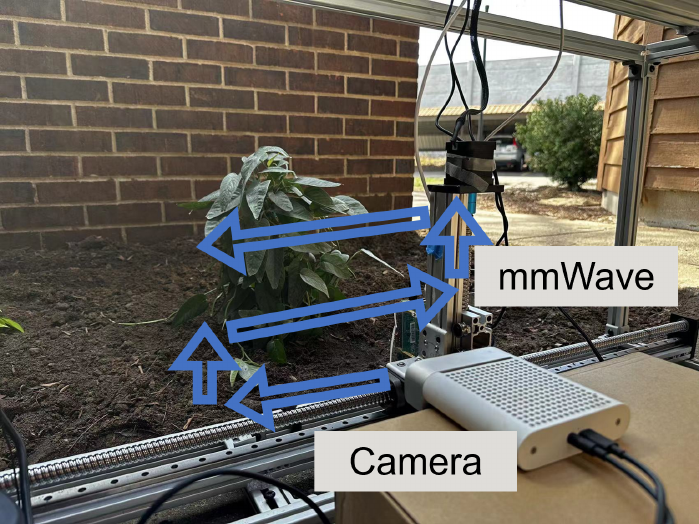}
       \caption{Outdoor Setup}
   \end{subfigure}
   \begin{subfigure}[b]{0.196\textwidth}
       \centering
       \includegraphics[width=\textwidth]{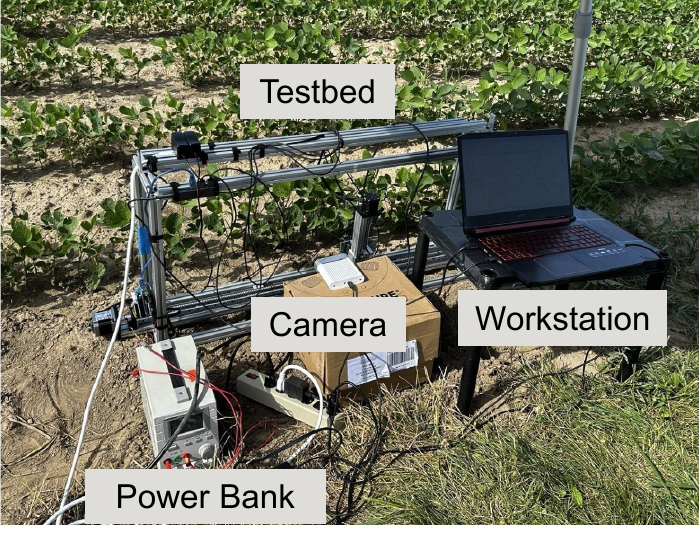}
       \caption{Soybean Field Setup}
   \end{subfigure}
   \begin{subfigure}[b]{0.196\textwidth}
       \centering
       \includegraphics[width=\textwidth]{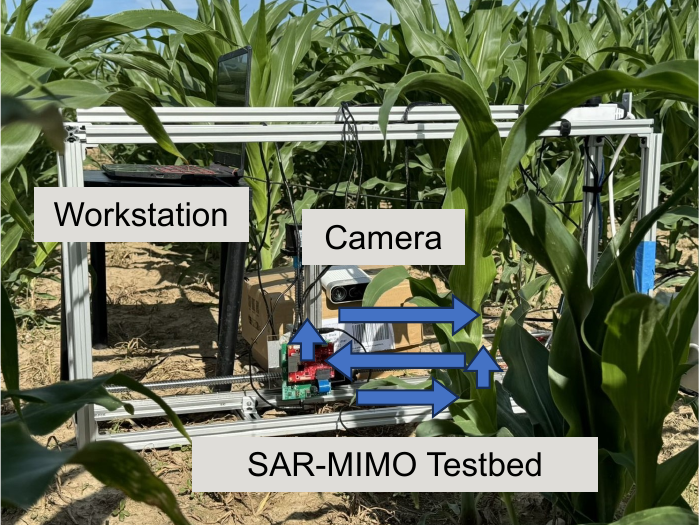}
       \caption{Corn Field Setup}
   \end{subfigure}
   \caption{\ours comprises three components: a mmWave radar, a two-axis scan testbed, and an RGB camera.}
   \label{fig:position}
   \Description{Setup of the \ours}
    \vspace{-2mm}
\end{figure*}

We developed a prototype of {\ours \footnote{The source design are available at \emph{https://github.com/liuyime2/MobiCom24-Hydra}.}}  As shown in Figure~\ref{fig:position}, the SAR Imaging system features a two-axis mechanical scanner designed to enable high-speed acquisition.
The scanner's design parameters are finely tuned for optimal plant analysis. It has a horizontal traversal capacity of $150$ mm and a vertical range of $100$ mm, aligning with the expected synthesis aperture size and the dimensions of our target plant.
The radar on the scanning mechanism is a Texas Instruments (TI) IWR1642 ~\cite{mmwave-radar}, operating within the 77 to 81 GHz frequency range.
This radar unit is essential for capturing the raw mmWave signals reflected from the targets under study.
Complementing the radar, a DCA 1000EVM ~\cite{mmwave-receiver} is employed for initial signal collection and processing, setting the stage for detailed imaging analysis.
In addition to the SAR Imaging system, our prototype includes a camera imaging component, leveraging an Azure Kinect camera ~\cite{kinect}, which is placed at the center of the scanning area.
This camera captures reflections in the visible light spectrum, thus providing a multimodal imaging perspective.
Integrating camera imaging with mmWave radar-based SAR imaging offers a holistic view of the target, significantly enhancing the analysis's depth and accuracy.
In evaluating the precision performance of our models, we've adopted the rigorous approach of applying 5-fold cross-validation, iterated ten times across different datasets.
This method is crucial for reducing the potential variability and bias that might arise from the dataset's diversity, ensuring that our performance metrics accurately reflect the model's capability under various conditions.

\subsection{System Preparation}

\begin{figure}[t]
   \begin{subfigure}[b]{0.25\textwidth}
       \centering
       \includegraphics[width=\textwidth]{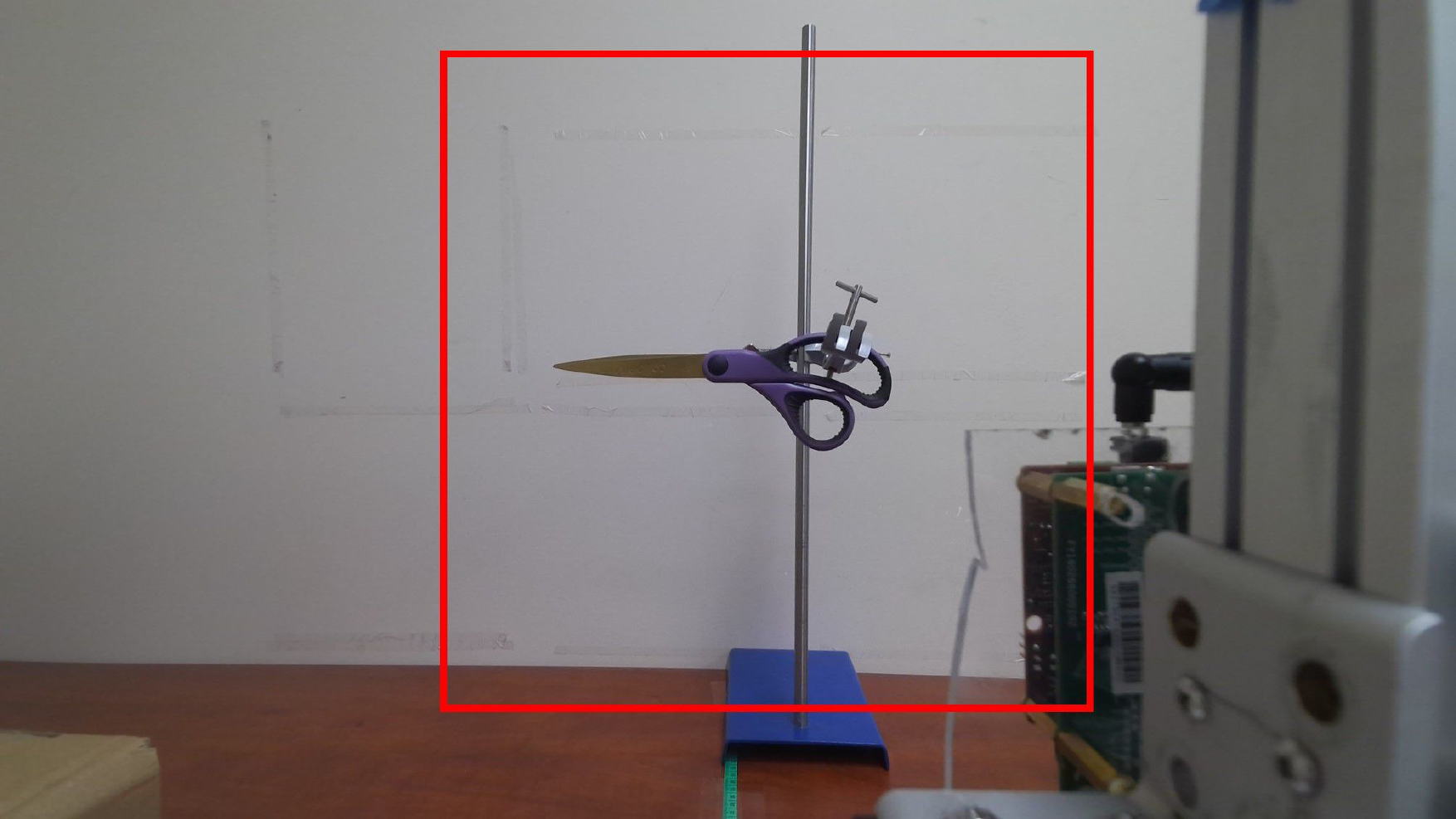} 
       \caption{Camera Image}
   \end{subfigure}
   \begin{subfigure}[b]{0.14\textwidth}
       \centering
       \includegraphics[width=\textwidth]{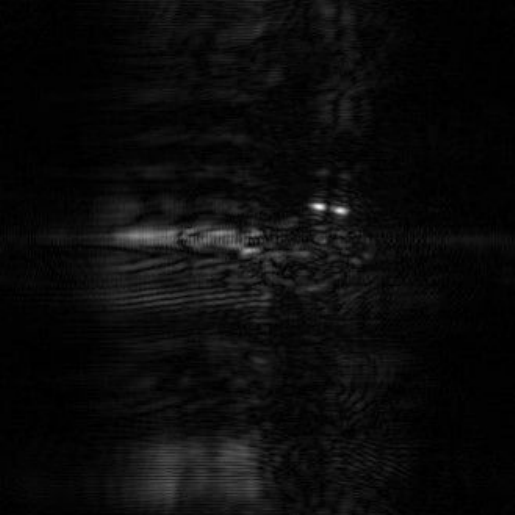}
       \caption{SAR Image}
       \label{fig:data-calibration-result}
   \end{subfigure}
   \caption{Calibration of the RGB camera and SAR sensor with the scissor. The red rectangle is the corresponding field of view for the MIMO sensor.}
   \label{fig:data-calibration}
   \Description{Data calibration setup}
   \vspace{-3mm}
\end{figure}

\noindent \textbf{Data Calibration.}  In the system design outlined in Section \ref{sec:single-depth-fe}, the successful fusion of SAR and RGB camera images hinges on aligning their FoV precisely. 
This alignment is crucial for integrating these images effectively in subsequent training processes.
Given the complexity of the match-up, calibration between SAR imaging and camera imaging systems becomes a significant task.
To address this problem, we employ a novel calibration strategy using a stable, depth-consistent object: the scissor.
This object is chosen because it can remain at a fixed depth, making it an ideal calibration target.
The calibration process is illustrated with the SAR imaging results of the scissor displayed in Figure \ref{fig:data-calibration}. 
We precisely match the RGB camera image, highlighted by the red block in Figure \ref{fig:data-calibration-result}.

\noindent \textbf{Feature Extractor.} 
\label{sec:feature-extraction}
In Section \ref{sec:single-depth-fe}, we detail a sophisticated feature extractor developed to process complex inputs comprising camera and SAR images, laying the groundwork for the subsequent Multi-Depth Leaf Wetness Detection module introduced in Section \ref{sec:multi-depth}.
We have explored the utility of pre-trained CNN models, utilizing a benchmark dataset ImageNet ~\cite{ImageNet2009}, renowned for its extensive range of image classes tailored for classification tasks.
Our investigation encompasses leading models like ResNet~\cite{ResNet2016}, VGG~\cite{VGG2014}, DenseNet~\cite{DenseNet2017}, and InceptionV3 ~\cite{Inception2015}, each acclaimed for their architectural innovations and exceptional performance in image classification challenges.
Our comparative analysis, presented in Table \ref{tab:model_structure}, reveals that ResNet surpasses its counterparts in accuracy for our specific application. 
The exceptional performance of ResNet is primarily attributed to its revolutionary architecture. It integrates residual learning to overcome the vanishing gradient problem~\cite{residual2016he}, a common challenge in DNN.
This design is particularly effective for processing the complex, high-dimensional data associated with SAR imaging, thereby enhancing leaf wetness's feature extraction and classification process amidst the diverse and complex plant foliage shapes.

\subsection{Data Collection}
\label{sec:data-collect}
\begin{figure}[!t]
   \centering
   \begin{subfigure}[b]{0.20\textwidth}
       \centering
       \includegraphics[width=\textwidth]{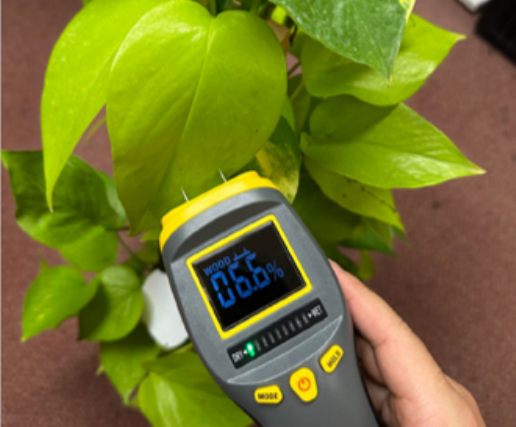}
       \caption{Dry Leaf}
   \end{subfigure}
   \hfill 
   \begin{subfigure}[b]{0.20\textwidth}
       \centering
       \includegraphics[width=\textwidth]{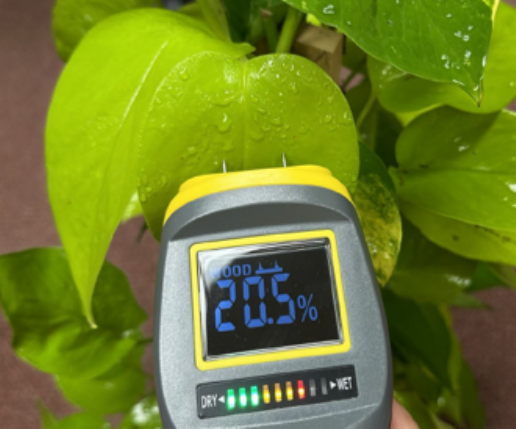}
       \caption{Wet Leaf}
   \end{subfigure}
   \caption{Using Moisture Meter for the ground truth Measurement.}
   \label{fig:ground_truth_sensor}
   \Description{Groundtruth sensor setup}
   \vspace{-2mm}
\end{figure}

\noindent \textbf{Ground truth.}
\rev{
Without a dedicated sensor to accurately track changes in LWD, we have adopted a novel approach to establishing reliable ground truth.
We employ a commercial moisture meter from General Tools~\cite{moisturemeter}, traditionally used for measuring wood surface wetness levels.
This device operates on the principle of electrical resistance measurement between two pins, as depicted in the accompanying Figure ~\ref{fig:ground_truth_sensor}.
Although not initially designed for leaf surface wetness detection, this instrument offers a viable alternative for our purposes.
By gently tapping the moisture meter's pins onto the leaf surface, we can observe variations in electrical resistance that correlate with the leaf's moisture content; higher water presence increases the readings.
This method allows us to measure the wetness level on the leaf surface directly.
Based on these resistance readings, we can effectively establish a threshold to differentiate between various leaf dryness and wetness stages.
}

\noindent \textbf{Baseline.}
In our experimental setup, we initially employed the commercial LWS, specifically the PHYTOS 31~\cite{PHYTOS31}, to monitor leaf wetness.
We use the PHYTOS 31 default threshold for the readings to determine whether the leaf is wet or dry.
We faced a significant challenge with the sensor's metal components, which enhanced backscatter and potentially affected the SAR imaging system's data quality. Additionally, these components obstructed light, creating a barrier for the RGB camera.
To mitigate these issues, we selected an analogous plant of similar size and placed it within the same environment as the monitored specimen.
To ensure comprehensive coverage and accurate assessment of leaf wetness across the plant, we positioned four different sensors, distributing them evenly to capture the entirety of the plant's foliage.
Given the inherent variation in baseline dryness resistance across different sensors, we focused on analyzing the average incremental change in resistance to determine leaf wetness.
This method bypasses the variability of individual sensor readings, leveraging the collective data from multiple sensors to provide a more reliable and consistent measure of LWD.
We also incorporate baseline comparisons for camera~\cite{patel2021strawberry} and mmLeaf~\cite{gan2023poster} technologies to ensure a comprehensive approach to wetness detection.
We selected ResNet-18 as the optimal model for camera-based wetness detection based on comparisons in Table \ref{tab:model_structure}. 
For the mmLeaf system, we followed the procedures in ~\cite{gan2023poster} to align our scanning and modeling with best practices for wetness detection.

\begin{table}[!t]
\caption{Comparison Methods for Detecting Wetness}
  \label{tab:model_structure}
  \begin{tabular}{ccc}
    \toprule
    Pre-trained Model & SAR Image Acc & Camera Acc \\
    \midrule
    \texttt{DenseNet} & $87.58\% \pm 6.10\%$  & $83.62\% \pm 1.97\%$ \\
    \texttt{VGG-16} & $85.50\% \pm 4.31\%$  & $82.42\% \pm 1.83\% $\\
    \texttt{InceptionV3} & $89.96\% \pm 5.69\%$ & $83.16\% \pm 2.13\%$ \\
    \texttt{ResNet-18} & $92.61\% \pm 3.43\%$ & $85.85\% \pm 1.31\%$ \\
    \bottomrule
    
  \end{tabular}
   \vspace{-2mm}
\end{table}

\noindent \textbf{Data Collection.}
\rev{Our indoor experiment involved a six-month data collection period from five diverse plant types, focusing on leaf size and orientation variations.
During this period, the plants exhibited different growth patterns, spacing, and distributions, increasing the dataset's diversity.
We used the same plants in outdoor experiments to assess performance in dynamic environments. 
Additionally, we conducted in-situ experiments on two fields totaling 3.67 acres, planted with soybeans and corn, to evaluate \ours in practical settings.
The experiments were performed under various scenarios, including different times of the day—early morning, afternoon, and evening—and diverse weather conditions, such as sunny, windy, and post-rain environments.
The experiment is set up with the plant 200-500 mm away from the mmWave radar and focusing on two extreme values: completely dry and saturated wet.
This approach maximizes the range of wetness levels, enabling us to distinguish wetness features effectively.
We compiled a dataset of approximately 674 pairs, including 536 pairs from indoor environments, 46 pairs from outdoor environments, and 92 pairs from real fields with dynamic environments.
Additionally, we developed a dataset for \ours to evaluate the performance of the LWD.
\ours monitor plants transitioning from fully saturated to completely dry indoors.
We sample the entire drying process 6 to 12 times.
Twenty groups are collected from varying plants and environmental conditions depending on the drying time.
}
This diverse collection helped highlight the differences in wetness features on leaf surfaces.
Leveraging the refined resolution capabilities of mmWave technology, we captured images at 1 mm depth intervals up to a distance of 300 mm, ensuring comprehensive coverage of plant information.
Additionally, to construct a precise timeline of the drying process, we monitored the changes in leaf wetness every 10 minutes, establishing a ground truth for drying duration.
Our evaluation also encompasses outdoor scenarios, covering sunny and windy conditions with temperatures between 10-17 Celsius degrees, spanning from morning to night, to consider the potential for dew formation. 
Given \ours's limitation of not operating in the rain, we simulated rainy conditions for the plant alone to assess the system's adaptability and performance under varied environmental influences.

\noindent \textbf{Baseline.}
In our experimental setup, we initially employed the commercial LWS PHYTOS 31~\cite{PHYTOS31}, to monitor leaf wetness.
We use the PHYTOS 31 default threshold for the readings to determine whether the leaf is wet or dry.
We faced a significant challenge with the sensor's metal components, which enhanced backscatter and potentially affected the SAR imaging system's data quality. Additionally, these components obstructed light, creating a barrier for the RGB camera.
To mitigate these issues, we selected an analogous plant of similar size and placed it within the same environment as the monitored specimen.
To ensure comprehensive coverage and accurate assessment of leaf wetness across the plant, we positioned four different sensors, distributing them evenly to capture the entirety of the plant's foliage.
Given the inherent variation in baseline dryness resistance across different sensors, we focused on analyzing the average incremental change in resistance to determine leaf wetness.
This method bypasses the variability of individual sensor readings, leveraging the collective data from multiple sensors to provide a more reliable and consistent measure of LWD.
We also incorporate baseline comparisons for camera~\cite{patel2021strawberry} and mmLeaf~\cite{gan2023poster} technologies to ensure a comprehensive approach to wetness detection.
We selected ResNet-18 as the optimal model for camera-based wetness detection based on comparisons in Table \ref{tab:model_structure}. 
For the mmLeaf system, we followed the procedures in ~\cite{gan2023poster} to align our scanning and modeling with best practices for wetness detection.

\section{EVALUATION}
This section details the performance evaluation of \ours through comprehensive over-the-air experiments, showcasing its capabilities in real-world scenarios.

\subsection{Overall Performance}
\label{sec:overall_performance}
\begin{figure}[!t]
    \vspace{4mm}
   \centering
   \begin{subfigure}[b]{0.24\textwidth}
       \centering\includegraphics[width=\textwidth]{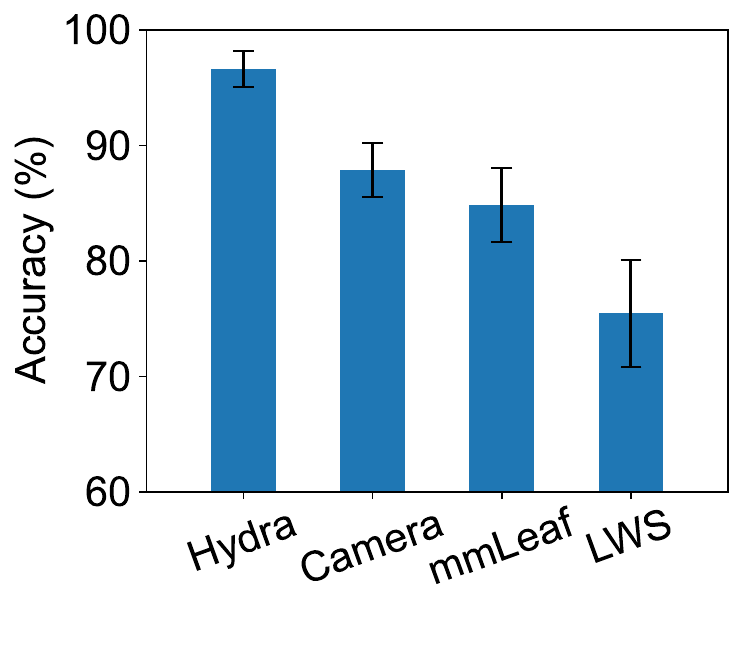}
       \captionsetup{skip=-2pt}
       \caption{Wetness Detection Acc}\label{fig:overall_acc}
   \end{subfigure}
   \hfill
   \begin{subfigure}[b]{0.23\textwidth}
       \centering\includegraphics[width=\textwidth]{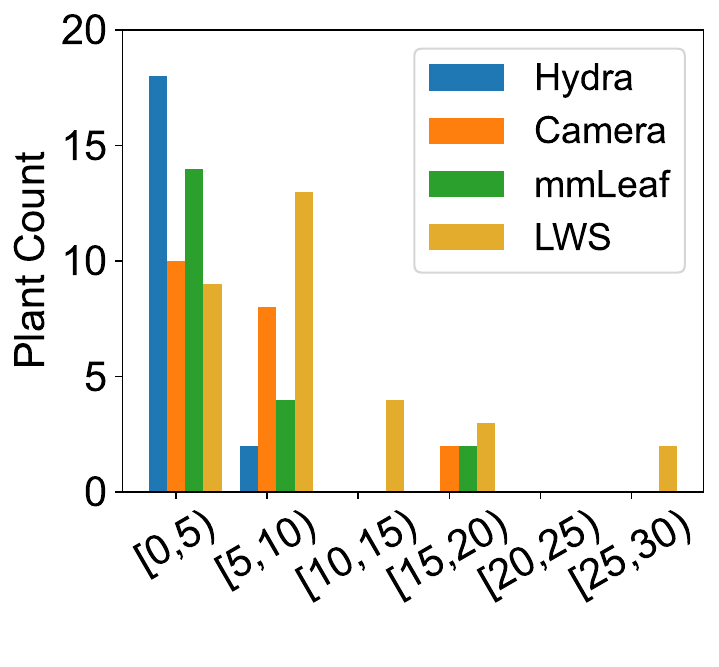}
       \captionsetup{skip=-2pt}
       \caption{Performance of LWD}\label{fig:overall_duration}
   \end{subfigure}
   \caption{\ours performance of detection LWD.}
   \label{fig:LWS_result}
   
   \Description{Overall result}
   \vspace{-2mm}   
   \end{figure}

In our evaluation of \ours, we focused on its precision in discerning various levels of leaf surface wetness.
Utilizing an extensive dataset, our system demonstrated an exceptional ability to identify, achieving a median precision rate of $96.21\% \pm 2.57\%$, as illustrated in Figure \ref{fig:overall_acc}.
This performance was compared against traditional methods, camera imaging, the mmLeaf system, and LWS systems, where \ours outperformed with accuracies of $87.8\% \pm 2.32\%, 83.84\% \pm 3.21\%$ and $75.25\% \pm 4.43\%$, respectively.

In the LWD analysis, a comparative test across 20 different data groups as detailed in Section \ref{sec:data-collect} pitted \ours against conventional methods: LWS, Camera, and mmLeaf systems.
The result shown in Figure \ref{fig:overall_duration} demonstrates \ours's accuracy, detecting LWD accurately within a 2-minute error margin in 18 out of 20 groups.
In comparison, the Camera system's performance varied, with eight groups having 5-10-minute errors and two groups showing 15-20-minute errors.
The mmLeaf system saw four groups with 5-10-minute errors and two with 15-20-minute errors.
LWS results were the most varied, with errors spanning from 5 minutes to 30 minutes across different groups, highlighting \ours's precision in detecting the LWD.

These results highlight the advancements \ours brings to the field by addressing the limitations of other modalities.
Camera, as discussed in Section \ref{sec:camera-preliminary}, lacks depth information, making it difficult to detect wetness when leaves overlap.
The mmLeaf system's effectiveness is compromised by a lack of environmental resilience, as noted in Section \ref{sec:mmWave-preliminary}.
LWS accuracy heavily depends on precise sensor placement.
The differences in weight and surface texture can affect wetness dynamics, leading to lower accuracy and higher variability, as stated in Section \ref{sec:lws-preliminary}.\ours's comprehensive approach combines depth information and surface texture analysis, overcoming these limitations.

\subsection{Single Depth Fusion Performance}
In \ours, we introduced a single depth fusion designed to merge data from camera image and SAR imaging, targeting accurately identifying leaf wetness features.
To evaluate the effectiveness of our single-depth fusion approach, we conducted a comparative study against established data fusion techniques, specifically early fusion and late fusion.
Early fusion combines input data at the model's onset, while late fusion integrates features before the classification stage.
Additionally, our analysis compares our unique data enhancement strategies, as detailed in Section \ref{sec:model_training}.
In our evaluation, we compared our single-depth fusion module, which integrates RGB camera and SAR imaging data, against traditional early and late fusion methods.
Our analysis, detailed in Figure \ref{fig:fusion-comparison}, highlighted that conventional methods struggle with the dimensional differences between modalities, limiting detection accuracy to $83.42\% \pm 1.84\%$ and $82.83\% \pm 1.47\%$ for early and late fusion, respectively.
In contrast, our fusion module significantly outperformed these, achieving an accuracy of $90.37\% \pm 2.24\%$.
Data augmentation techniques, including removing and adjusting RGB camera images, enhanced our module's precision to $93.02\% \pm 1.79\%$.
The results demonstrate our module's ability to utilize multimodal data effectively.
Data enhancement leads to more stable results, effectively allowing the model to uncover deeper features within both SAR and RGB images, which can reduce the impact of overfitting.

\subsection{Scan Motion Performance}
\begin{figure}[!t]
   \centering
   \begin{subfigure}[b]{0.225\textwidth}
       \centering\includegraphics[width=\textwidth]{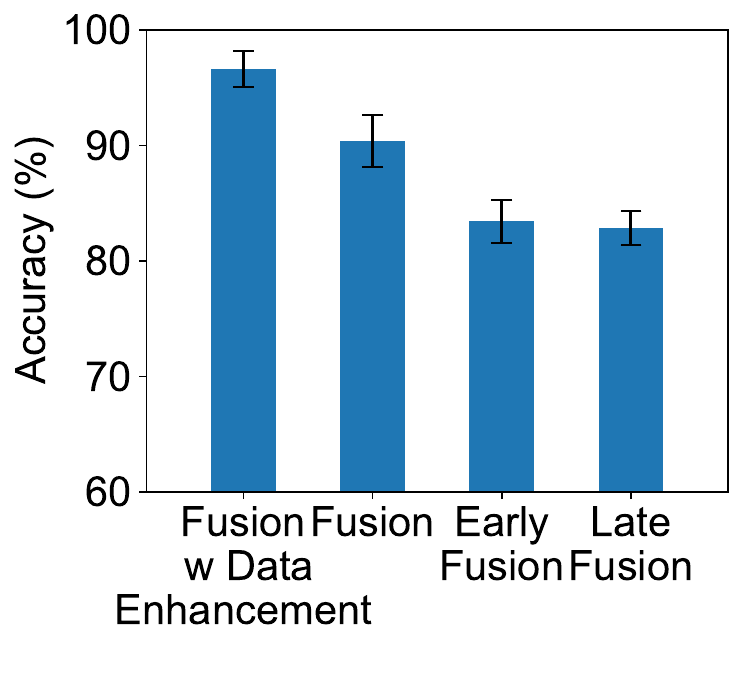}
    \captionsetup{skip=-2pt}
       \caption{Fusion Performance}
   \label{fig:fusion-comparison}
   \end{subfigure}
   \hfill
   \begin{subfigure}[b]{0.245\textwidth}
       \centering\includegraphics[width=\textwidth]{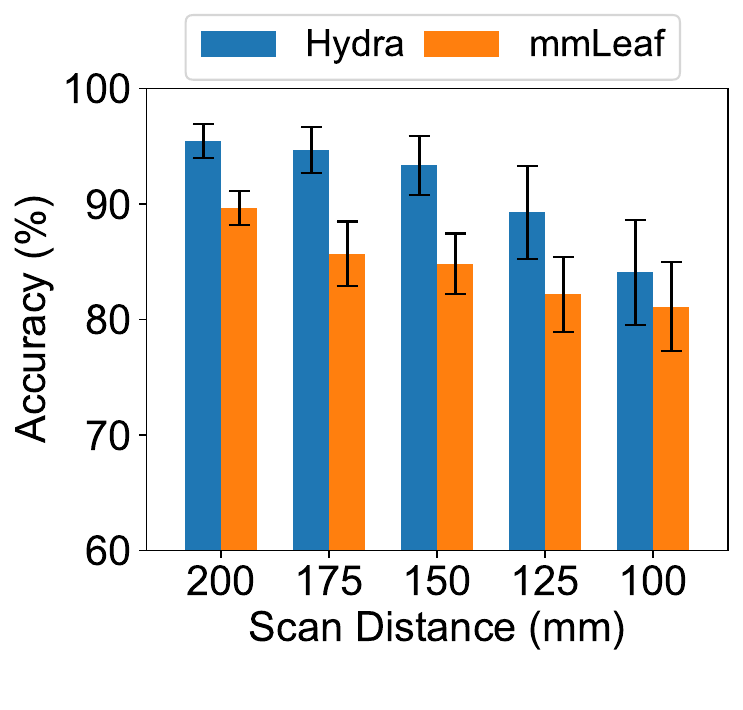}
    \captionsetup{skip=-2pt}
       \caption{Scan Dist. Performance}
       \label{fig:scan_distance}
   \end{subfigure}
   \caption{Performance of Multi-Modal fusion and SAR system's performance across scan distances }
   \vspace{-4mm}
   \label{fig:scan-fov}
    
   \Description{Fusion result and efficiency result}   
\end{figure}

As mentioned in Section \ref{sec:intro}, efficiency is crucial for large farm applications.
Increased efficiency allows farms to monitor more plants with \ours, leading to more accurate disease control.
This section evaluates scanning duration and FoV to optimize our system's efficacy without compromising accuracy.
In our evaluation, we trained the model using a single type of plant fixed at 200 mm from the mmWave sensor under optimal lighting, focusing exclusively on assessing the effects of scan distance modifications and comparing them with mmLeaf.
Through a series of experiments, we systematically adjusted the scanning field from a comprehensive 200 mm, ensuring entire plant coverage, down to a focused 100 mm, centering on the plant's core.
As shown in Figure \ref{fig:scan_distance}, for \ours, our findings establish a direct correlation between the width of the scanning field and detection accuracy.
With a maximal field of 200mm, the system achieved its highest accuracy when reducing the FoV, which led to a steady decline in accuracy.
For the scan distance longer or equal to 150 mm, with the advantage of the multi-modality, the accuracy is above 90\%, which is $95.43\% \pm 1.47\%, 94.68\% \pm 1.97\% $ and $93.38\% \pm 2.56\%$ for 200mm, 175mm and 150mm respectfully.
For the scan distance lower than 150mm, the accuracy reduces dramatically with a higher variance of $89.28\% \pm 4.05\%$ and $84.1\% \pm 4.53\%$. The degradation in performance can be attributed to a decrease in image resolution.
Additionally, a shorter scan distance causes \ours to miss critical parts necessary for accurate detection. For the mmLeaf, the accuracy drops rapidly from 200 mm to 100 mm; the accuracy is $89.66\% \pm 1.46\%, 85.71\% \pm 2.79\%, 84.83\% \pm 2.63\%, 82.18\% \pm 3.27\%, 81.13\% \pm 3.83\%$.
This trend underscores that reducing the scan distance leads to a smaller synthetic aperture and narrower FoV, which diminishes the detection of wetness features on leaves.
However, leveraging multimodal features can compensate for this loss, ensuring the SAR, compared to mmLeaf, has a shorter efficiency of 25\% and retains high accuracy by enhancing the overall picture of leaf wetness through integrating diverse data sources.

\begin{figure*}[!t]
   \centering
   \begin{subfigure}[b]{0.23\textwidth}
       \centering\includegraphics[width=\textwidth]{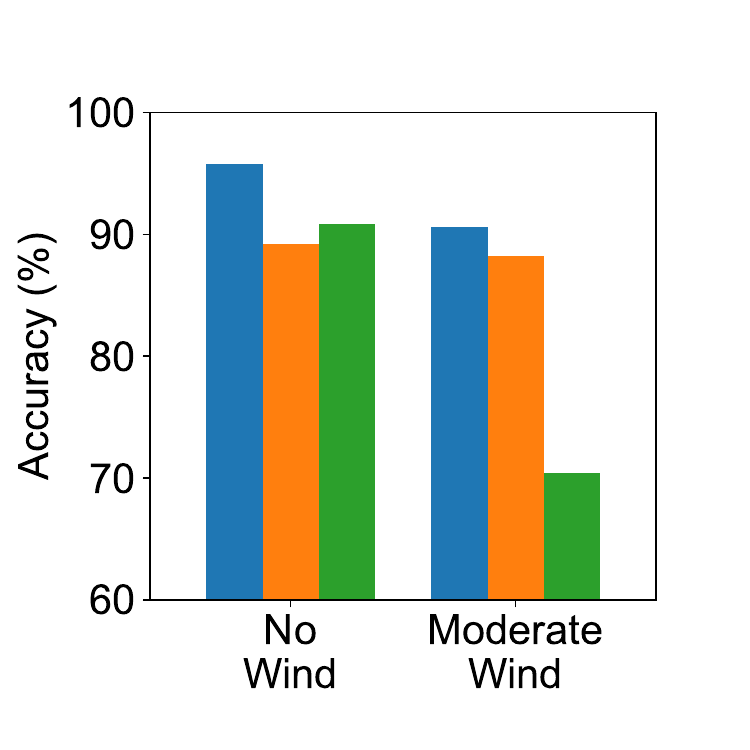}
   \caption{Windy}\label{fig:wind-anlaysis} 
   \end{subfigure}
   \begin{subfigure}[b]{0.23\textwidth}
       \centering\includegraphics[width=\textwidth]{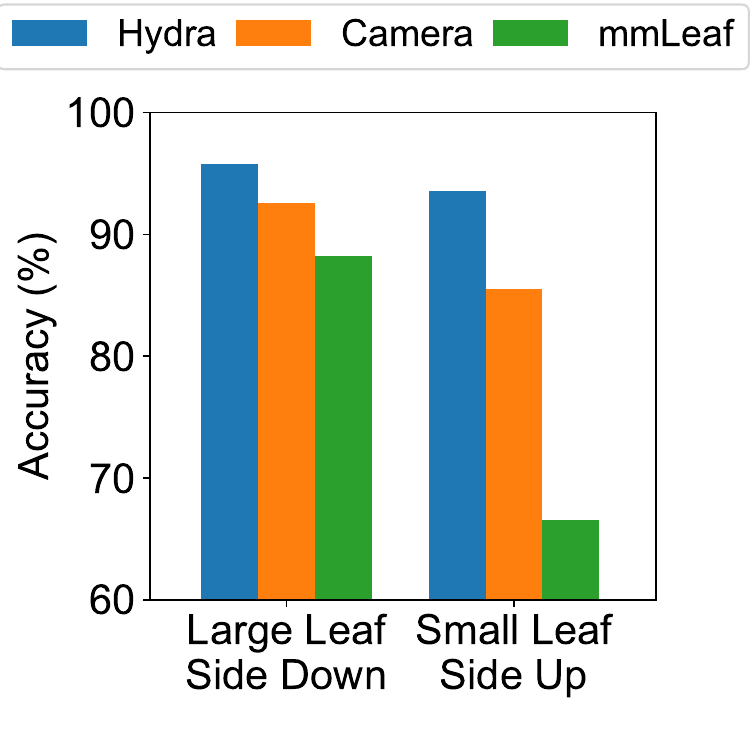}
   \caption{Different Leaf}\label{fig:leaf-analysis}
   \end{subfigure}
   \begin{subfigure}[b]{0.23\textwidth}
       \centering\includegraphics[width=\textwidth]{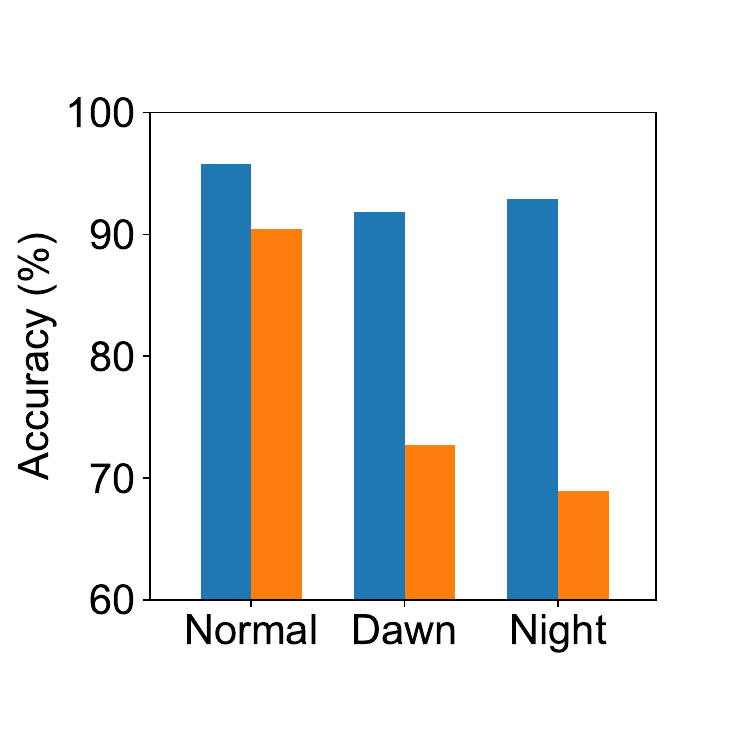}
   \caption{Lighting Condition}\label{fig:light-analysis}
   \end{subfigure}
   \begin{subfigure}[b]{0.23\textwidth}
       \centering\includegraphics[width=\textwidth]{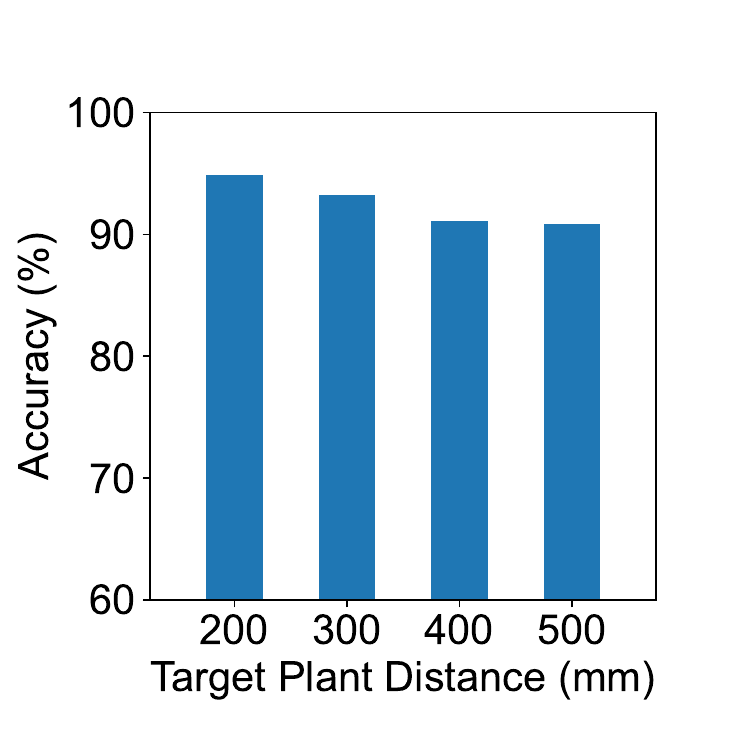}
   \caption{Plant Distance}\label{fig:distance_change}
   \end{subfigure}
   \label{fig:different-env}
   \caption{Performance for \ours application in a different real-world scenario.}
   \Description{Component wise result}
\end{figure*}

\begin{figure}[!t]
   \centering
   \begin{subfigure}[b]{0.12\textwidth}
       \centering\includegraphics[width=\textwidth]{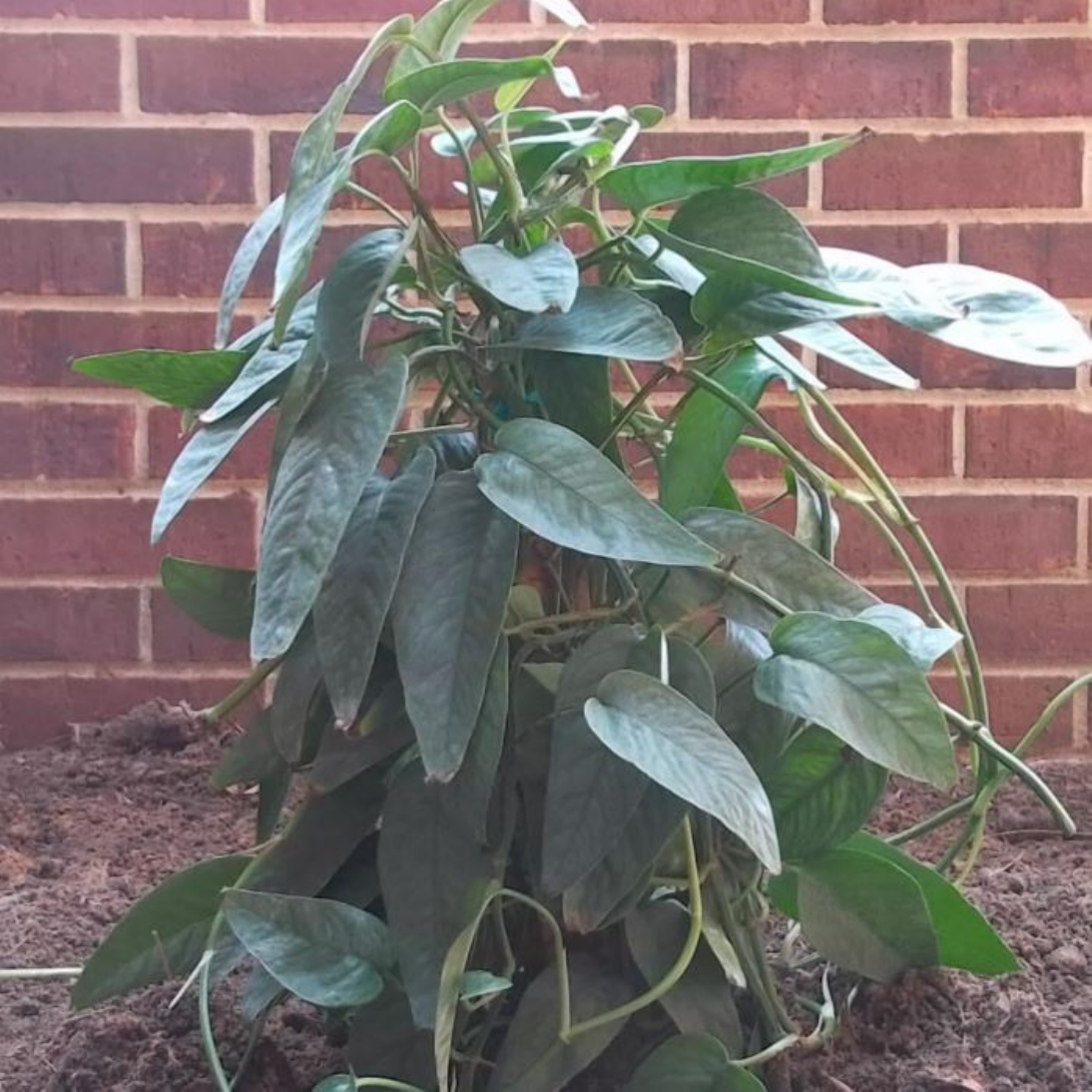}
   \end{subfigure}
   \begin{subfigure}[b]{0.12\textwidth}
       \centering\includegraphics[width=\textwidth]{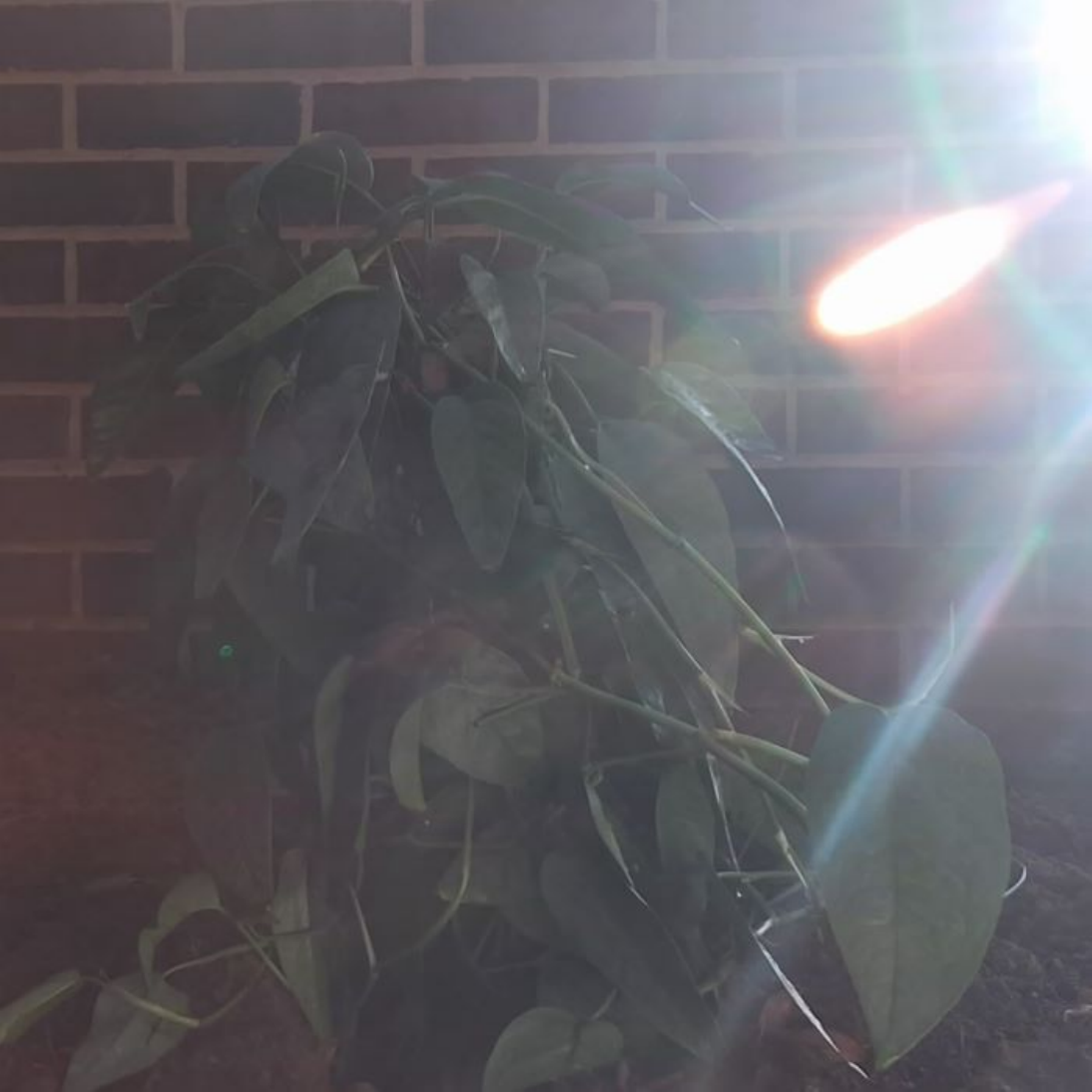}
   \end{subfigure}
   \begin{subfigure}[b]{0.12\textwidth}
       \centering\includegraphics[width=\textwidth]{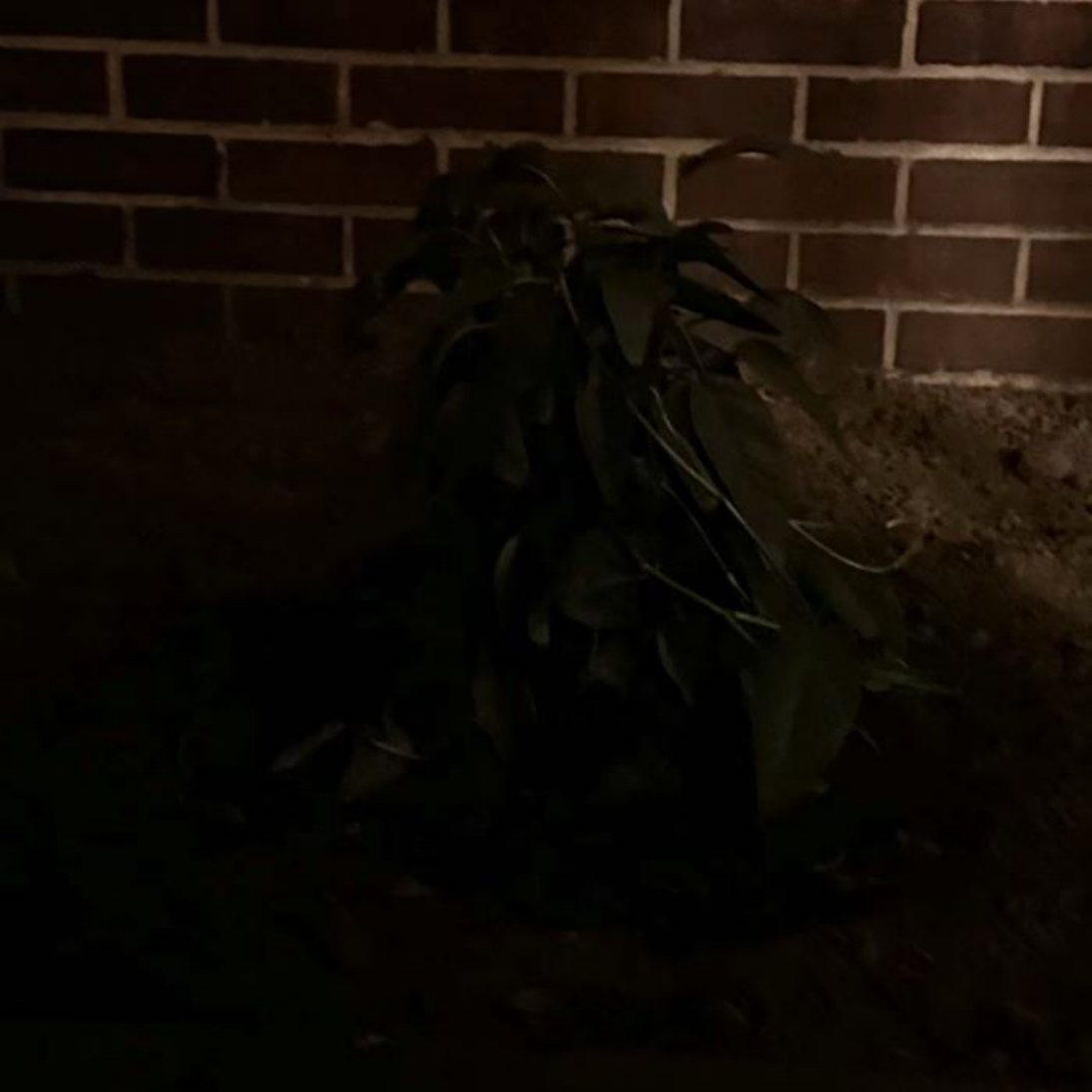}
   \end{subfigure}
   \begin{subfigure}[b]{0.12\textwidth}
       \centering\includegraphics[width=\textwidth]{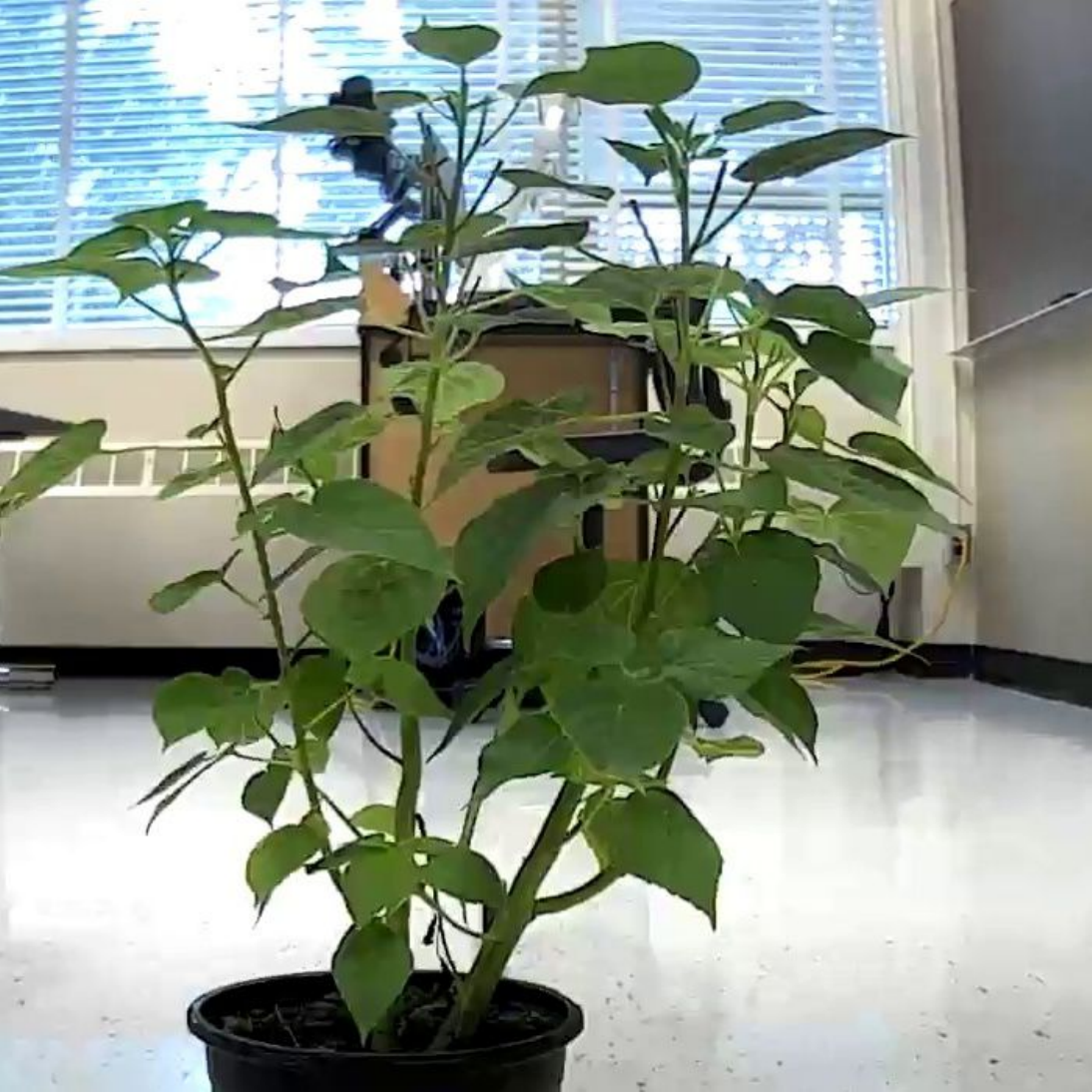}
   \end{subfigure}
   \begin{subfigure}[b]{0.12\textwidth}
       \centering\includegraphics[width=\textwidth]{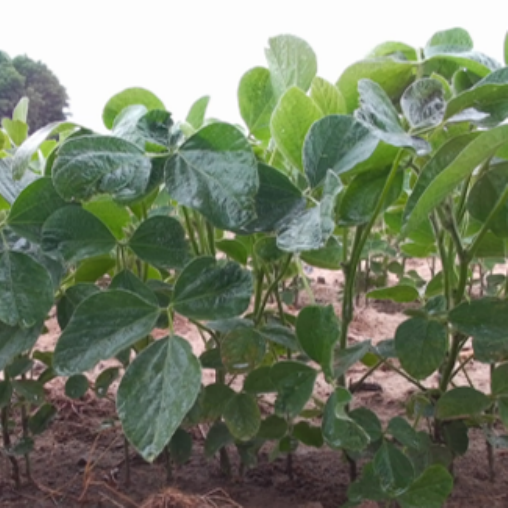}
   \end{subfigure}
   \begin{subfigure}[b]{0.12\textwidth}
       \centering\includegraphics[width=\textwidth]{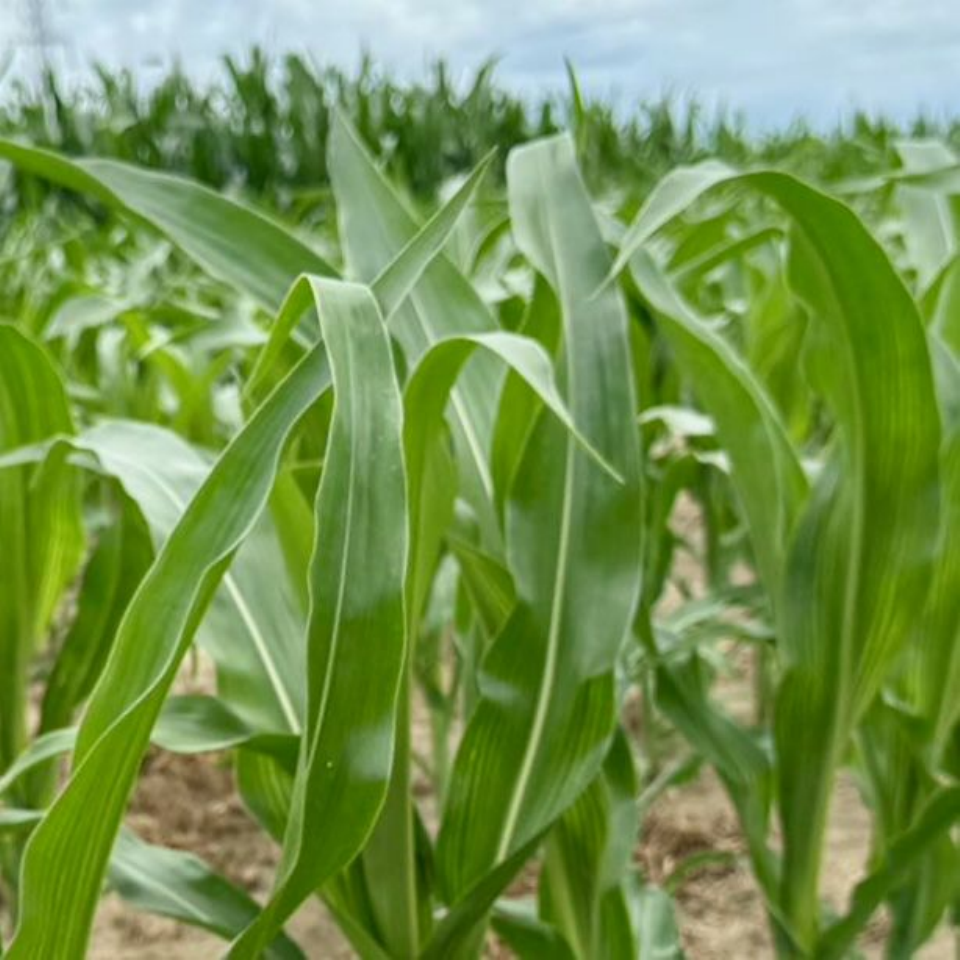}
   \end{subfigure}
   \caption{\ours application in different scenarios.}
   \label{fig:different-light}
   \Description{Outdoor Result}
   \vspace{-4mm}
\end{figure}

\begin{figure*}[t]
   \centering
   \begin{subfigure}[b]{0.23\textwidth}
       \centering\includegraphics[width=\textwidth]{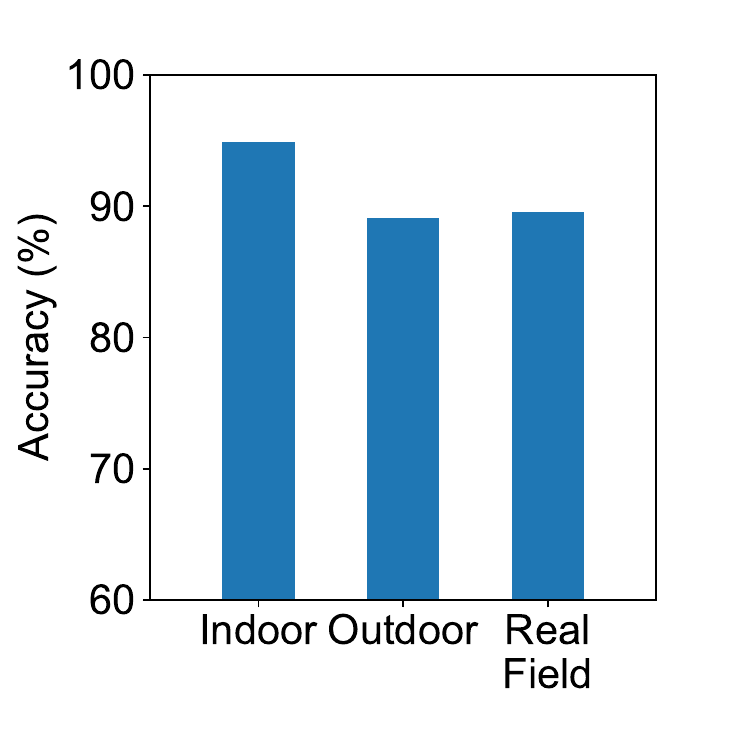}
       \captionsetup{skip=-2pt}
   \caption{Environment}\label{fig:indoor-outdoor}                   
   \end{subfigure}                                                       
   \begin{subfigure}[b]{0.23\textwidth}
       \centering\includegraphics[width=\textwidth]{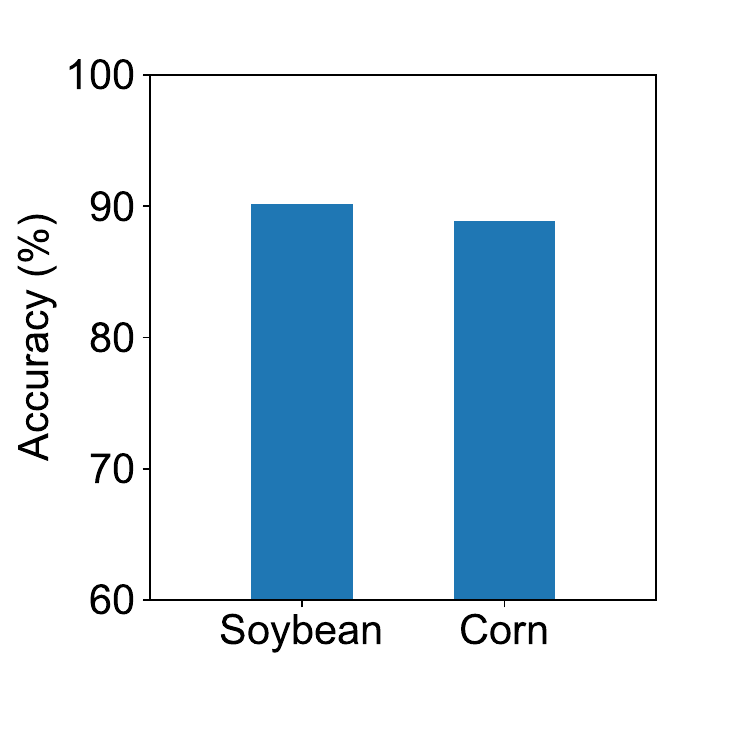}
       \captionsetup{skip=-2pt}
   \caption{Crop}\label{fig:farm-plant}
   \end{subfigure}  
   \begin{subfigure}[b]{0.23\textwidth}
       \centering\includegraphics[width=\textwidth]{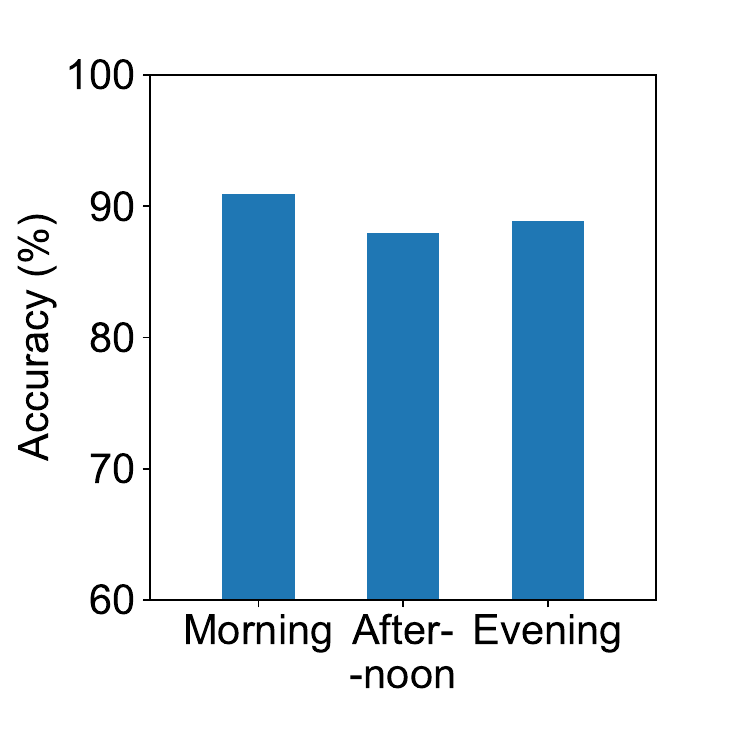}
       \captionsetup{skip=-2pt}
   \caption{Time of Day}\label{fig:farm-time}
   \end{subfigure}
   \begin{subfigure}[b]{0.23\textwidth}
       \centering\includegraphics[width=\textwidth]{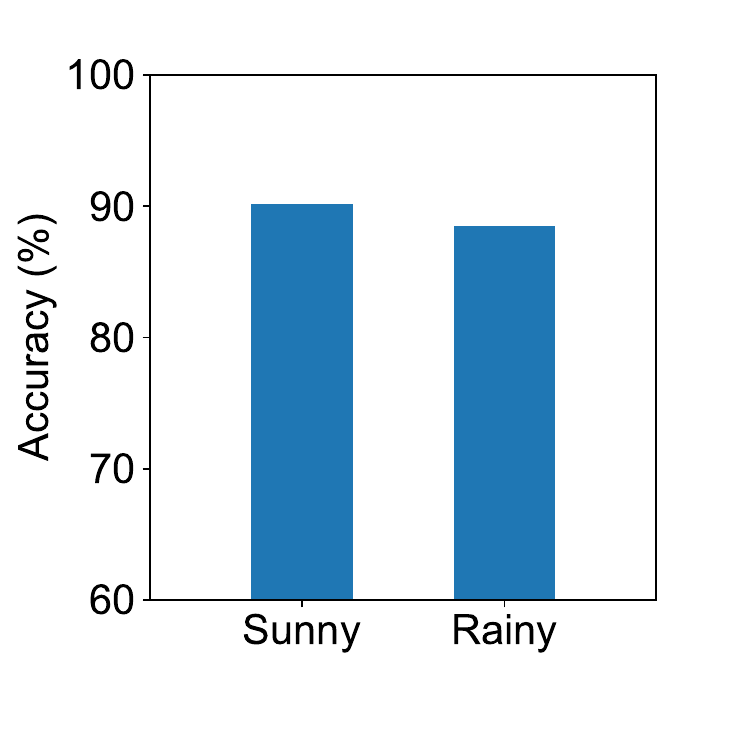}
       \captionsetup{skip=-2pt}
   \caption{Weather}\label{fig:farm-weather}
   \end{subfigure}
   \label{fig:real-farm}
   \caption{Performance for \ours application in a different real-world scenario.}
   \label{fig:real-analysis}
   \Description{Component wise result}
   \vspace{-2mm}
\end{figure*}

\subsection{Application in Different Environment}
We evaluated sensors' performance in the indoor environment and simulated real farm environments as shown in Figure \ref{fig:different-light}, focusing on how environmental factors impact the accuracy of wetness detection technologies relative to the baseline.
Our study involves three sensors: \ours, a standard camera, and mmLeaf. 
We examined their precision in detecting leaf wetness while considering the analyzed environmental variables.

\rev{We evaluate the wind environment under a controlled indoor environment using a fan to simulate the impact of the wind.}
The result is illustrated in Figure \ref{fig:wind-anlaysis}.
Without wind, accuracy rates were high across all devices: $95.76\%$ for \ours, $90.85\%$ for the camera, and $89.2\%$ for mmLeaf.
The introduction of moderate wind saw a decline, particularly affecting \ours and mmLeaf due to their reliance on SAR imaging, where movement disrupts phase alignment, leading to errors.
Despite this, \ours maintained a leading precision rate of  $90.62\%$.
The camera's quicker imaging speed has little influence on achieving $88.2\%$.
But mmLeaf dropped to $70.43\%$, showcasing the wind has significantly influenced the SAR imaging process.
However, \ours maintains its robustness in wind scenarios thanks to its multi-modal features and a training process designed for resilience.

\rev{We evaluate \ours performance for the different types of leaves in the indoor environment, which focuses on the leaf orientation and size.} For larger leaves positioned leaf-side down, \ours, camera, and mmLeaf showcased exceptional accuracy, registering at $95.76\%, 92.61\%$ and $88.22\%$ as shown in Figure \ref{fig:leaf-analysis}.
\ours precision experienced a minor decline to $93.53\%$ when the system was tasked with evaluating smaller leaves oriented leaf-side up and a dramatic degradation for camera and mmLeaf of the precision $85.5\%, 66.57\%$.
The precision shows that \ours is robustness to the plant with different morphology. 
The precision of \ours demonstrates its robust ability to adapt to plants with varying morphologies, ensuring accurate wetness detection across different plant types.

\rev{We evaluate the impact of different lighting conditions from normal to dawn to night in the outdoor simulated real farm environment.}
The analysis, summarized in Figure \ref{fig:light-analysis}, demonstrates that lighting significantly influences camera-based detection, with accuracy dropping in low-light conditions.
In regular light, the camera can reach an accuracy of $90.42 \%$, and at Dawn and night, it will drop to $72.73 \%$ and $68.97 \% $.
However, \ours exhibited minimal impact from varying light conditions, underscoring its comprehensive multimodal analysis capability and resilience in diverse environmental settings.
The accuracy would drop a little but still keep the accurate precision, which the normal light, dawn time, and night time are $95.63\%, 91.87\%, 92.92\%$, respectively.

Further, evaluate \ours system assessed its performance across different distances from a target plant in an indoor setting as shown in Figure \ref{fig:distance_change}.
The highest accuracy, $94.86\%$, was achieved when the plant was 200mm away.
As the distance increased to 300mm, 400mm, and 500mm, accuracy slightly decreased to $93.23\%$, $90.08\%$, and $91.94\%$, respectively, indicating a relationship between increased distance and diminished mmWave resolution.
The results suggest that while \ours with the multi-modal maintains robust performance across varying distances, optimizing placement relative to the target can enhance detection precision.

\rev{
\subsection{Application in Real Field}
We conducted real-field experiments in soybean and corn fields, including early morning, dawn, evening, sunny, and rainy weather.
Our goal was to examine \ours precision in dynamic, practical environments.

We first simulated practical conditions outdoors and then tested in soybeans and corn fields.
The results, shown in Figure \ref{fig:indoor-outdoor}, indicate that the simulated outdoor environment and real farm conditions achieved similar accuracies of 89.13\% and 89.6\%, respectively.
Although these are slightly lower than the indoor setup accuracy of 94.64\%, they still demonstrate superior performance.

Soybeans and corn are two distinct types of crops that differ significantly in leaf size and arrangement.
Soybeans have smaller, denser leaves, while corn features more extensive, sparsely arranged ones.
The results for \ours are shown in Figure \ref{fig:farm-plant}.
We achieved a high accuracy rate of 88.89\% for corn and 90.2\% for soybeans.
Evaluating diverse leaf structures with high accuracy demonstrates \ours robustness in detecting wetness across various crop types.

The environment in a real farm is inherently complex due to its ever-changing lighting and weather conditions.
Lighting varies significantly throughout the day, from soft morning light to intense afternoon sunlight to dim evening light.
These changes affect the visibility and reflectivity of leaves, impacting the accuracy of \ours.
Figure \ref{fig:farm-time} shows an accuracy of 90.91\% in the morning, 88\% at dawn, and 88.89\% in the evening.
The relatively lower accuracy during dawn suggests that direct sunlight might mislead the camera.
In the morning, \ours benefits from optimal multimodality conditions, while in the evening, it relies mainly on mmWave information.
Additionally, weather conditions such as sunny and rainy periods further complicate the environment.
We observed an accuracy of 90.2\% during sunny conditions and 88.46\% during rainy conditions, as shown in Figure \ref{fig:farm-weather}.
Results from evaluations at different times of day and various weather conditions demonstrate \ours's robustness in real farm settings.

}

\section{RELATED WORK}
\noindent \textbf{Leaf Water Content.}  
Innovations in wireless technology for monitoring plant moisture include methods like RFID, backscatter, and PCR technologies, primarily designed to measure Leaf Water Content (LWC)~\cite{daskalakis2018uw, dey2020paper, hoog202260}. Additionally, technologies such as LoRa~\cite{prism24dong, demeter24dong, ct24dong, is22dong}, ultrasound~\cite{echo23gen}, and infrared sensing~\cite{pyro24zeng} offer non-destructive and non-invasive means to accurately assess LWC.

\noindent \textbf{mmWave Multimodality.}  
Recent advancements in mmWave and multimodal technologies have enhanced health diagnostics and wireless communication. Studies~\cite{EffectiveGaitAbnormalityDetection, mDSPCGR} demonstrate improved gait recognition and health monitoring accuracy through mmWave imaging and micro-Doppler fusion. Additionally, multimodal features enhance model training, object detection, depth estimation, and mmWave beam selection in wireless networks~\cite{MultiDimensionalInformationFusion, RIDERS, wang2022detection}.

\noindent \textbf{3D Classification.}  
The \ours model uses 3D data analysis to tackle complex classification challenges, enhancing precision in identifying intricate patterns beyond 2D capabilities. In healthcare, \ours improves diagnostic accuracy in tumor identification and anatomical mapping, supporting personalized treatment plans. It has been applied to detect moldy corn~\cite{corn2024zhang} and analyze sleep postures for better healthcare outcomes~\cite{sleepposition2024liu}.
\rev{\section{PRACTICAL DEPLOYMENT DISCUSSION}
Deploying \ours in agriculture enhances monitoring and disease control, leveraging advanced technology to optimize practices and increase crop yields in large farms and greenhouses.

Drones are increasingly used for precision agriculture because they cover large areas efficiently.
For large farms, \ours can be mounted on drones to autonomously navigate predefined routes, providing real-time data collection across extensive fields.
This approach is supported by studies on drone applications in crop monitoring and pesticide spraying~\cite{HAFEEZ2023192, lachgar2023unmanned}.

In greenhouses or dense fields where drones may struggle with leaf wetness detection due to space constraints and thick foliage, \ours can utilize rail systems for continuous monitoring~\cite{pathmudi2023systematic}.
These rail-mounted systems efficiently navigate greenhouses, offering targeted analysis by sampling selected plants.
This method allows precise data collection and real-time environmental adjustments, optimizing plant health and growth in confined or densely planted areas.
 }

\section{CONCLUSION} 
This paper introduces the development, deployment, and comprehensive evaluation of \ours, an advanced system for precise leaf wetness detection.
\ours tackles the critical challenge of accurately assessing leaf wetness, a significant factor in agricultural productivity and disease management.
We introduced \ours to incorporate and analyze data from multiple sources, including innovative SAR imaging and an RGB camera.
We conducted extensive experiments to evaluate \ours in practical environments.
Our experiments included various plants and environmental factors. 
The results show that \ours consistently surpasses existing methodologies in accuracy and reliability.
\ours demonstrated a remarkable 96\% accuracy in determining leaf wetness across diverse settings.
When deployed on a real farm, \ours consistently maintained around 90\% accuracy.

\section*{Acknowledgement}
\label{sec:ack}
\noindent
We sincerely thank the anonymous reviewers and our shepherd for their valuable feedback. This work was partially supported by NSF CAREER Award 2338976.

\bibliographystyle{ACM-Reference-Format}
\bibliography{main}

\end{document}